\documentclass[pdflatex,sn-basic]{sn-jnl}

\usepackage{graphicx}%
\usepackage{multirow}%
\usepackage{amsmath,amssymb,amsfonts}%
\usepackage{amsthm}%
\usepackage{mathrsfs}%
\usepackage[title]{appendix}%
\usepackage{xcolor}%
\usepackage{textcomp}%
\usepackage{manyfoot}%
\usepackage{booktabs}%
\usepackage{algorithm}%
\usepackage{algorithmicx}%
\usepackage{algpseudocode}%
\usepackage{listings}%

\usepackage[capitalise]{cleveref}

\DeclareMathOperator{\Tanh}{Tanh}
\DeclareMathOperator{\LeakyReLU}{LeakyReLU}

\DeclareMathOperator{\WhPred}{\textit{wh\_pred}}
\DeclareMathOperator{\Label}{\textit{label}}
\DeclareMathOperator*{\ArgMax}{arg\,max}
\DeclareMathOperator{\Similarity}{sim}
\DeclareMathOperator{\Length}{\textit{len}}
\DeclareMathOperator{\NL}{NL}

\usepackage{xcolor}

\usepackage{pifont}
\definecolor{darkgreen}{rgb}{0.0, 0.5, 0.0}  
\newcommand{\darkgreencheck}{{\color{darkgreen}\ding{51}}}  
\definecolor{darkred}{rgb}{0.6, 0.0, 0.0}  
\newcommand{\darkredcross}{{\color{darkred}\ding{55}}}  

\usepackage{soul}

\usepackage{multirow}

\usepackage{stackengine}
\setstackEOL{\\}
\makeatletter
\def\BState{\State\hskip-\ALG@thistlm}
\makeatother

\usepackage{array}

\usepackage{booktabs}

\theoremstyle{thmstyleone}%

\theoremstyle{thmstyletwo}%

\theoremstyle{thmstylethree}%

\begin{document}

\title[GNN2R: Weakly-Supervised Rationale-Providing Question Answering over Knowledge Graphs]{GNN2R: Weakly-Supervised Rationale-Providing Question Answering over Knowledge Graphs}

\author*[1,2]{\fnm{Ruijie} \sur{Wang}\textsuperscript{\href{https://orcid.org/0000-0002-0581-6709}{[0000-0002-0581-6709]}}}\email{ruijie.wang@uzh.ch}

\author[3]{\fnm{Luca} \sur{Rossetto}\textsuperscript{\href{https://orcid.org/0000-0002-5389-9465}{[0000-0002-5389-9465]}}}\email{luca.rossetto@dcu.ie}

\author[4,5]{\fnm{Michael} \sur{Cochez}\textsuperscript{\href{https://orcid.org/0000-0001-5726-4638}{[0000-0001-5726-4638]}}}\email{michael.cochez@vu.nl}

\author[1]{\fnm{Abraham} \sur{Bernstein}\textsuperscript{\href{https://orcid.org/0000-0002-0128-4602}{[0000-0002-0128-4602]}}}\email{bernstein@ifi.uzh.ch}

\affil*[1]{\orgname{University of Zurich}, \orgaddress{\city{Zurich}, \country{Switzerland}}}

\affil*[2]{\orgname{SIB Swiss Institute of Bioinformatics}, \orgaddress{\city{Zurich}, \country{Switzerland}}}

\affil[3]{\orgname{School of Computing, Dublin City University}, \orgaddress{\city{Dublin}, \country{Ireland}}}

\affil[4]{\orgname{Ellis Institute Finland}%
}

\affil[5]{\orgname{Åbo Akademi University}, \orgaddress{\city{Turku}, \country{Finland}}}

\abstract{
Despite the rapid progress of large language models (LLMs), knowledge graph-based question answering (KGQA) remains essential for producing verifiable and hallucination-resistant answers in many real-world settings where answer trustworthiness and computational efficiency are highly valued.
However, most existing KGQA methods provide only final answers in the form of KG entities.
Without explicit explanations---ideally in the form of intermediate reasoning process over relevant KG triples, the QA results are difficult to inspect and interpret.
Moreover, this limitation prevents the rich and verifiable knowledge encoded in KGs, which is a key advantage of KGQA over LLMs, from being fully leveraged.
However, addressing this issue remains highly challenging due to the lack of annotated intermediate reasoning process and the requirement of high efficiency in KGQA.
In this paper, we propose a novel \textbf{G}raph \textbf{N}eural \textbf{N}etwork-based \textbf{Two}-Step \textbf{R}easoning method (\textbf{GNN2R}) that can efficiently retrieve both final answers and corresponding reasoning subgraphs as verifiable rationales, using only weak supervision from widely-available final answer annotations.
We extensively evaluated GNN2R and demonstrated that GNN2R substantially outperforms existing state-of-the-art KGQA methods in terms of effectiveness, efficiency, and the quality of generated explanations.
The complete code and pre-trained models are available at \url{https://github.com/ruijie-wang-uzh/GNN2R}.}

\keywords{Knowledge Graph, Question Answering, Graph Neural Network, Language Model}

\maketitle

\section{Introduction}

\begin{figure}[t]
    \centering
    \includegraphics[trim=0pt 9pt 0pt 10pt, clip, width=0.6\linewidth]{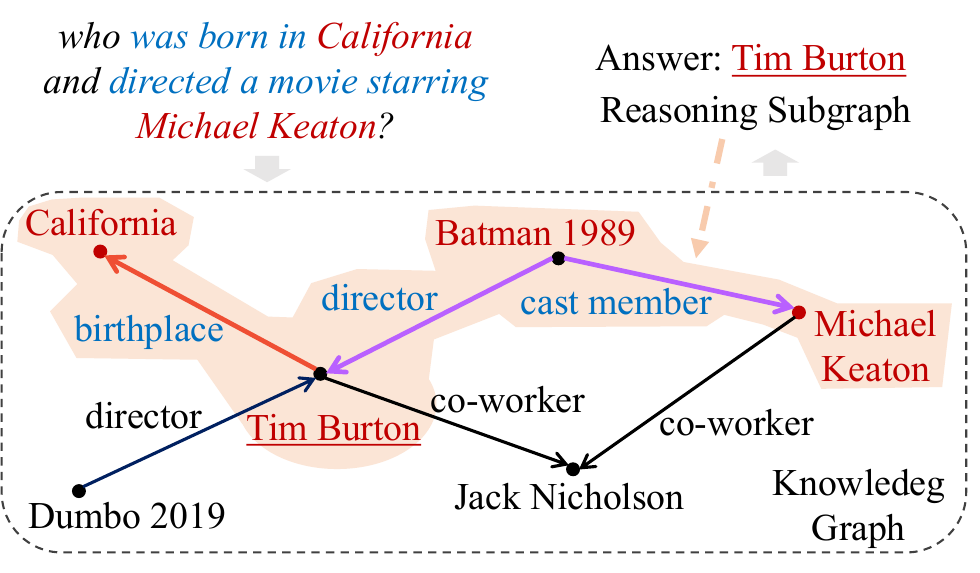}
    \caption{A KGQA example, where question-relevant entities and relations are highlighted in red and blue.
    The reasoning subgraph is distinguished by a light-orange background, featuring two reasoning paths in orange and purple.} 
    \label{fig:introduction}
\end{figure}

In recent years, knowledge graphs (KGs) have been prominently used to organize and provide knowledge for question answering (QA) tasks.~\citep{DBLP:journals/datamine/ChongLL25,DBLP:journals/datamine/JinZYTYL23}
An example is given in \cref{fig:introduction}, where the question ``\textit{who was born in California and directed a movie starring Michael Keaton}'' can be answered based on the given KG with \texttt{Tim Burton}.
However, it is difficult to examine and understand this answer for normal users without a computer science background, given the technical complexity of navigating the question-relevant parts within a large KG.
We propose to provide users with a \emph{reasoning subgraph} consisting of triples from the KG that elucidate how the answer satisfies the semantic constraints specified in the question.
As highlighted in the figure, the reasoning subgraph of this question consists of two paths:
[(\texttt{Tim Burton}, \texttt{birthplace}, \texttt{California})], which affirms that Tim Burton was born in California, and [(\texttt{Tim Burton}, \texttt{director\textsuperscript{-1}}, \texttt{Batman 1989}),
\footnote{We use the superscript \textsuperscript{-1} to denote the inverse of a relation.}
(\texttt{Batman 1989}, \texttt{cast member}, \texttt{Michael Keaton})], which affirms that Tim Burton directed Batman 1989 starring Michael Keaton.
Hence, \emph{reasoning subgraphs provide users with a \textbf{rationale} for the final answers}.
Also, the intermediate information in reasoning subgraphs, e.g., the alluded movie in this example, could be useful and important for fully understanding the answer.

The generation of reasoning subgraphs is challenging due to the following issues:
First, KGQA is expected to function online and provide responses promptly.
Our method needs to maintain high efficiency even after incorporating the retrieval of reasoning subgraphs, especially when compared to existing methods that omit this process.
Second, there is a lack of annotations of reasoning subgraphs in KGQA datasets that can be utilized as supervision for this task.
Particularly considering complex questions that involve multi-hop reasoning in KGs, it is challenging to train a model for reasoning subgraph generation based on only annotations of final answers.
It is worth mentioning that the recent large language models (LLMs)~\citep{DBLP:journals/tist/ChangWWWYZCYWWYZCYYX24} have demonstrated the potential of automatically annotating reasoning subgraphs.
However, several challenging issues remain to be addressed, including quality control of the automatic annotations, considering the potential inconsistency, bias, and noise that LLMs may introduce.
Also, KGs often organize queried information differently from how it is expressed in questions, which is commonly referred to as the semantic gap~\citep{DBLP:journals/kais/DiefenbachLSM18} between KGs and natural language.
Substantial efforts are required to mitigate this issue when using LLMs to annotate reasoning subgraphs for natural language questions.

To achieve rationale-providing QA while handling the above challenges, we propose a novel \textbf{G}raph \textbf{N}eural \textbf{N}etwork-based \textbf{Two}-step \textbf{R}easoning model, called \textbf{GNN2R}, which answers questions in two steps:
First, a novel graph neural network (GNN) is proposed to represent KG entities and given questions in joint embedding spaces where questions are close to their answers.
This step efficiently prunes the search space of candidate answers and, consequently, prunes the search space of reasoning subgraphs.
Second, we handle the lack of reasoning subgraph annotations by jointly utilizing the existing structures of KGs and the strong language understanding capability of a pre-trained language model (LM).
Simple yet effective algorithms are proposed for the subgraph extraction and the weakly-supervised fine-tuning of LM in this step.

\noindent We summarize our contributions in this paper in the following:
\begin{itemize}
    \item We introduce a novel and challenging task setup that addresses a long-neglected gap in KGQA: retrieving both final answers and reasoning subgraphs for the examination and understanding of final answers.
    \item We propose a novel QA model---GNN2R that can efficiently generate both final answers and reasoning subgraphs with only weak supervision.
    \item We propose a novel graph neural network (GNN) that is highly tailored to the reasoning of complex questions in KGQA and can effectively prune the search space of final answers and reasoning subgraphs.
    \item We propose a weakly-supervised method for the extraction of reasoning subgraphs based on the joint utilization of existing KG structures and the language understanding capability of the pre-trained language model.
    \item We conducted extensive experiments demonstrating the superior performance of our model regarding effectiveness, efficiency, and the quality of generated reasoning subgraphs compared to state-of-the-art (SOTA) methods.
\end{itemize}

\section{Related Work}
\label{sec:related_work}

\begin{table*}[t]
\centering
\caption{Summary of gaps of existing KGQA methods in comparison with our method. Please refer to \cref{sec:related_work} for detailed discussions and references.}
\label{table:limitations}
\begin{tabular}{>{\centering\arraybackslash}m{3.5cm} >{\centering\arraybackslash}m{1.8cm} >{\centering\arraybackslash}m{2cm} >{\centering\arraybackslash}m{2cm} >{\centering\arraybackslash}m{1.8cm}}
\toprule
 &  Weak Supervision &  Possibility of Returning Explanation &  Detailed Reasoning Explanation &  SOTA Performance \\ \midrule
Query Construction-based Methods   & \darkredcross & \darkgreencheck & \darkgreencheck & \darkredcross \\ \midrule
Information Retrieval-based Methods & \darkgreencheck & \darkredcross & \darkredcross & \darkredcross \\ \midrule
Explicit Reasoning-based Methods & \darkredcross & \darkgreencheck & \darkgreencheck & \darkredcross \\ \midrule
Implicit Reasoning-based Methods & \darkgreencheck & \darkgreencheck & \darkredcross & \darkredcross \\ \midrule
Our Method                           & \darkgreencheck & \darkgreencheck & \darkgreencheck & \darkgreencheck \\ \bottomrule
\end{tabular}
\end{table*}

In this section, we review and compare with several major categories of related KGQA methods, and discuss the limitations of large language models (LLMs) in the context of KGQA.

\emph{Query construction-based methods}~\citep{DBLP:journals/corr/abs-2204-08554,DBLP:conf/acl/CaoS0LY0L0X22,DBLP:journals/tkde/ChenLQWW23,DBLP:journals/datamine/ChongLL25,DBLP:conf/semweb/0003ZRRB23} translate given questions into logical queries, e.g., SPARQL~\citep{DBLP:journals/tods/PerezAG09}, that can be executed over KGs to obtain answers.
By instantiating the constructed logical queries, this type of method can provide explicit reasoning subgraphs as answer explanations.
However, recent methods of this type, including case-based~\citep{DBLP:journals/corr/abs-2204-08554}, transfer learning-based~\citep{DBLP:conf/acl/CaoS0LY0L0X22}, and supervised learning-based~\citep{DBLP:journals/tkde/ChenLQWW23}, heavily rely on existing corpora or annotations of logical queries, making them inapplicable to the task of this paper, where only weak supervision (i.e., annotations of final answers) is available for training.

\emph{Information retrieval-based methods}~\citep{DBLP:conf/emnlp/SunDZMSC18,DBLP:conf/acl/SaxenaTT20,DBLP:conf/emnlp/YanLWZDZWX21,DBLP:journals/datamine/JinZYTYL23} directly retrieve answers from KGs via parametric ranking or classification of KG entities based on their contextual representations, e.g., KG embeddings. 
These methods are usually highly efficient and effective with only weak supervision.
However, these methods lack interpretability and cannot provide answer explanations or explicit reasoning subgraphs.

\emph{Explicit reasoning-based methods}~\citep{DBLP:conf/iclr/DasDZVDKSM18,DBLP:conf/wsdm/QiuWJZ20} aim to conduct explicit step-wise reasoning over KG triples, e.g., reinforcement learning-based exploration~\citep{DBLP:conf/iclr/DasDZVDKSM18}, to search for answer entities starting from topic entities of given questions.
Their explicit reasoning trajectory in the KG can be used as answer explanations and reasoning subgraphs.
However, the performance of these methods is usually limited when only weak supervision is available.

\emph{Implicit reasoning-based methods}~\citep{DBLP:conf/emnlp/ShiC0LZ21,DBLP:conf/naacl/0091SJ22,DBLP:conf/coling/ZhouHZ18} perform fuzzy reasoning by computing confidence scores or numerical labels for KG entities and triples.
These methods are effective with only weak supervision.
Moreover, their computed confidence scores and labels can be used as a form of answer explanation.
However, these methods are unable to provide explicit reasoning subgraphs.

The above-mentioned limitations of existing KGQA methods are summarized in \cref{table:limitations}.
In this work, we aim to propose a method that is effective with weak supervision and capable of providing answer explanations---particularly explicit reasoning subgraphs.
As demonstrated in experiments, our method fulfills these requirements and outperforms the current KGQA methods with state-of-the-art (SOTA) performance regarding QA effectiveness, run-time efficiency, and quality of generated reasoning subgraphs.

Recently, the application of LLMs in KGQA and conversational agents has attracted wide attention \citep{DBLP:journals/corr/abs-2302-06466,DBLP:conf/semweb/TanMLLHCQ23,DBLP:conf/naacl/SunXZLD24,DBLP:journals/datamine/MeemRH24}.
However, there are still several vital issues to address.
First, given the implicit nature of the knowledge in LLMs and the commonly-observed hallucination issue~\citep{DBLP:journals/corr/abs-2309-16459}, it is difficult to verify the validity and completeness of answers from LLMs.
In contrast, our method can return explicit and verifiable rationales behind final answers.
Second, compared with current KGQA methods, LLMs consume significantly more computational resources and are often prohibitively expensive to adapt when training is needed to incorporate updated data or private internal data.
In our experiments, training our model requires only a single NVIDIA GeForce RTX 2080 Ti GPU, making it several orders of magnitude more efficient in terms of computation costs compared to LLMs.
Third, as \citet{DBLP:conf/semweb/TanMLLHCQ23} highlighted, it is often tedious and difficult to extract concrete final answers from LLMs' textual responses of varying length, where answers may be expressed as aliases, acronyms, synonyms, or even in a different language.
Fourth, as \citet{DBLP:conf/naacl/SunXZLD24} showed, recent powerful LLMs, including GPT-4\footnote{\url{https://openai.com/index/gpt-4-research/}} and LLaMA (33B)~\citep{DBLP:journals/corr/abs-2302-13971}, still have difficulty answering simple (one-hop) factual questions, which is often considered a solved task with current KGQA methods.

\section{Problem Definition}
\label{subsection:problem_definition}

We denote a KG as $\mathcal{G} = (E, R, T)$, where $E$, $R$, and $T$ respectively refer to the entity set, relation set, and triple set of $\mathcal{G}$.
A triple $(e_h, r, e_t)\in T$ consists of a head entity $e_h \in E$, a relation $r \in R$, and a tail entity $e_t \in E$.
A natural language question posed over $\mathcal{G}$ is denoted as $q$. 
The answer to $q$ is a set of entities in $E$, denoted as $A_q \subseteq E$.
The reasoning subgraph of $q$, denoted as $\mathcal{G}_q$, is a subgraph of $\mathcal{G}$ that consists of triples linking answers in $A_q$ with topic entities mentioned in $q$ in the way that satisfies the semantic constraints specified in $q$.
In this paper, only a limited set of question-final answer pairs are available for training.
The task is to train a QA model that can retrieve both final answers and reasoning subgraphs for new and unseen questions.

\section{Method: The GNN2R QA Model}

GNN2R answers a given question in two steps: \textbf{a graph neural network (GNN)-based coarse reasoning step} and \textbf{a language model (LM)-based explicit reasoning step}.
In the first step (cf. \cref{subsection:step_1}), a GNN is trained to represent KG entities and the question in a joint embedding space, where the question is close to its answers.
We collect candidate answers via nearest neighbor search.
In the second step (cf. \cref{subsection:step_2}), we extract candidate reasoning subgraphs that link the candidate answers to topic entities mentioned in the given question.
Then, we fine-tune a pre-trained Sentence Transformer~\citep{DBLP:conf/emnlp/ReimersG19} to select the final reasoning subgraph that is the most semantically consistent with the given question.

\subsection{Step-I: GNN-based Coarse Reasoning}
\label{subsection:step_1}

\begin{figure}[t]
    \centering
    \includegraphics[trim=5pt 8pt 0pt 0pt, clip, width=0.6\textwidth]{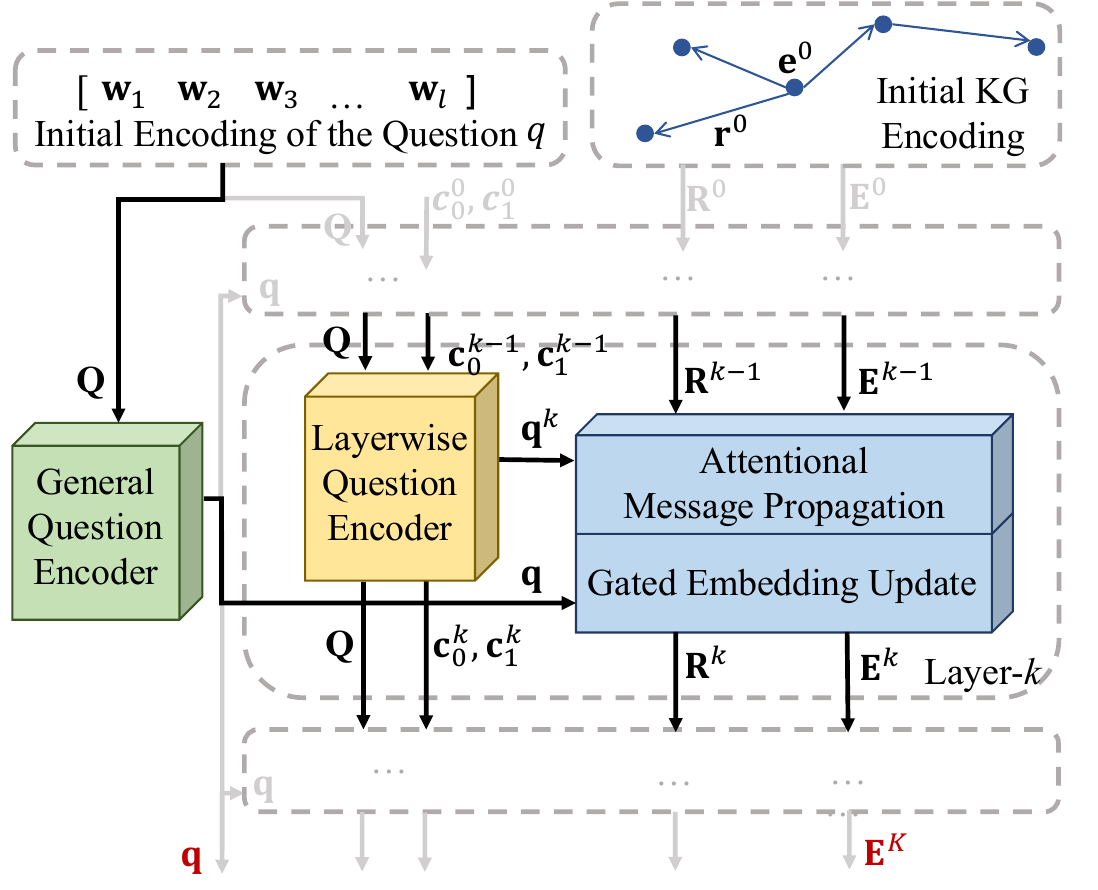}
    \caption{The GNN-based encoding of the given question and KG in Step-I.} 
    \label{fig:step_1}
\end{figure}

In the following, we introduce the four modules of the proposed GNN model, which are also presented as colored boxes in \cref{fig:step_1}.
We assume there are $K$ layers in the GNN and only depict the detailed architecture of the \textit{k}-th layer.

\textbf{The general question encoder} is implemented as a bidirectional single-layer gated recurrent unit (GRU) network~\citep{DBLP:conf/emnlp/ChoMGBBSB14} that computes the mean of the final hidden states from the two directions as the general question embedding $\mathbf{q} \in \mathbb{R}^{d_K}$, where $d_K$ denotes the embedding dimension.
The input of this module is $\mathbf{Q} =[\mathbf{w}_1, \mathbf{w}_2, ..., \mathbf{w}_{l}]$, which is the encoding of the given question $q$ based on a pre-trained BERT~\citep{DBLP:conf/naacl/DevlinCLT19} model.\footnote{\url{https://huggingface.co/bert-base-uncased}}

\textbf{The layerwise question encoder} provides layerwise reference information about the given question for the following attentional message propagation module.
For complex questions, the reference information may have different foci in different GNN layers.
This module is also implemented as a bidirectional single-layer GRU network with $\mathbf{Q}$ as input. 
The mean of its two final hidden states ($\mathbf{c}^{k}_{0}$ and $\mathbf{c}^{k}_{1}$) is used as the reference embedding of the \textit{k}-th layer: $\mathbf{q}^k \in \mathbb{R}^{d_k}$.
As we hope to consider the history of reference embeddings across different layers, the final hidden states from the previous layer ($\mathbf{c}^{k-1}_{0}$ and $\mathbf{c}^{k-1}_{1}$) are used as the initial hidden states of the current layer.
As for the first layer, zero vectors are used as initial hidden states.

\textbf{The attentional message propagation module} computes messages that KG entities receive to update their embeddings in different layers.
The input of this module in the $k$-th layer is the entity embedding matrix $\mathbf{E}^{k-1} \in \mathbb{R}^{|E| \times d_{k-1}}$ and the relation embedding matrix $\mathbf{R}^{k-1} \in \mathbb{R}^{|R| \times d_{k-1}}$ from the previous layer.
$|E|$ and $|R|$ are the number of entities and the number of relations in the KG.
In the first layer, analogous to the general question encoder, we employ BERT and GRU to encode relations based on their labels.
The embeddings of topic entities are initialized by the above general question embedding $\mathbf{q}$, while all other entities are initialized with a random vector that is considered as model parameters.

In this module, for each entity $e \in E$, if there is a relation $r_{j} \in R$ that links $e$ to a neighboring entity $e_i \in E$ regardless of the direction of the relation, i.e., $(e, r_j, e_i) \in T$ or $(e_i, r_j, e) \in T$, the entity $e$ receives a message $\mathbf{m}^k_{ij}$:
\begin{equation}
    \mathbf{m}^k_{ij} = \Tanh \left( \mathbf{W}^k_m [\mathbf{e}^{k-1}_i , \mathbf{r}^{k-1}_{j}] + \mathbf{b}^k_m \right),
    \label{equ:individual_message}
\end{equation}
where $\mathbf{W}^k_m \in \mathbb{R}^{d_k \times 2 d_{k-1}}$ and $\mathbf{b}^k_m \in \mathbb{R}^{d_k}$ are parameters of this module, and $[ \mathbf{e}^{k-1}_i , \mathbf{r}^{k-1}_{j} ]$ denotes the concatenation of the embeddings of $e_i$ and $r_{j}$ from the previous layer.

We denote the set of all the messages that $e$ receives as $M^k = \{\mathbf{m}^k_{ij}\}$.
Then, we employ the attention mechanism to compute an attentional aggregation of $M^k$.
Specifically, the attention score for each message $\mathbf{m}^k_{ij}\in M^k$ is computed as:
\begin{equation}
    s^k_{ij} = \mathbf{a}^k \LeakyReLU \left( \mathbf{W}^k_a [\mathbf{m}^k_{ij}, \mathbf{q}^k] + \mathbf{b}^k_a \right),
    \label{equ:individual_message_score}
\end{equation}
where $\mathbf{q}^k$ is the stepwise reference embedding, $\mathbf{a}^k \in \mathbb{R}^{1 \times d_k}$, $\mathbf{W}^k_a \in \mathbb{R}^{d_k \times 2 d_k}$, and $\mathbf{b}^k_a \in \mathbb{R}^{d_k}$ are parameters. 
Then, based on the Softmax function, we compute an attention weight $w_{ij}^k$ for each message $\mathbf{m}^k_{ij}$.
Finally, the attentional aggregation of all the messages can be formulated as follows:
\begin{equation}
    \mathbf{m}^k = \sum_{\mathbf{m}^k_{ij} \in M^k} w_{ij}^k \mathbf{m}^k_{ij}.
    \label{equ:aggregated_message}
\end{equation}

\textbf{The gated embedding update module} includes an attentional gate that determines to what extent the embedding of $e$ should be updated with the aggregated message $\mathbf{m}^k$.
This module utilizes the final question embedding $\mathbf{q}$ as the reference to compute two attention scores $s^k_m$ and $s^k_e$ for $\mathbf{m}^k$ and the current entity embedding $\mathbf{e}^{k-1}$, respectively:
\begin{equation}
    s^k_m = \mathbf{a}^k_u  \LeakyReLU \left( \mathbf{W}^k_u \mathbf{W}^k_{t1} [\mathbf{m}^k, \mathbf{q}] + \mathbf{b}^k_u \right),
    \label{equ:aggregated_message_score}
\end{equation}
\begin{equation}
    s^k_e = \mathbf{a}^k_u \LeakyReLU \left( \mathbf{W}^k_u \mathbf{W}^k_{t2} [\mathbf{e}^{k-1}, \mathbf{q}] + \mathbf{b}^k_u \right),
    \label{equ:previous_embedding_score}
\end{equation}
where $\mathbf{a}^k_u \in \mathbb{R}^{1 \times d_k}$, $\mathbf{W}^k_u \in \mathbb{R}^{d_k \times 2 d_k}$, $\mathbf{W}^k_{t1} \in \mathbb{R}^{2 d_k \times (d_k + d_K)}$, $\mathbf{W}^k_{t2} \in \mathbb{R}^{2 d_k \times (d_{k-1} + d_K)}$, $\mathbf{b}^k_u \in \mathbb{R}^{d_k}$ are parameters of this module.
Please note that $\mathbf{W}^k_{t1}$ and $\mathbf{W}^k_{t2}$ are just for dimension conversion and can be fixed to identity matrices when $d_k = d_{k-1} = d_K$.
Finally, assuming the Softmax-based attention weights computed for $s^k_m$ and $s^k_e$ as $w^k_m$ and $w^k_e$, the embedding of $e$ is updated as follows:
\begin{equation}
    \mathbf{e}^k = w^k_m \mathbf{m}^k + w^k_e \mathbf{e}^{k-1}.
    \label{equ:embedding_update}
\end{equation}

We perform the above computations for each entity in the KG to obtain the updated entity embedding matrix $\mathbf{E}^k$.
As for the relation embedding matrix, if the dimension of the \textit{k}-th layer is changed, i.e., $d_k \neq d_{k-1}$, a linear transformation can be applied to $\mathbf{R}^{k-1}$.
Otherwise, the relation embeddings can be directly passed to the next layer.

\textbf{Training and Inference.} 
For a training question $q$ with the answer set $A_q \subseteq E$, we randomly sample pairs of answer and non-answer entities from the Cartesian product of $A_q$ and $E \setminus A_q$.
The training loss of $q$ is formulated as follows:
\begin{equation}
    \label{equ:gnn_loss}
    loss(q) = \sum_{\left(e, e'\right) \in A_q \times \left(E \setminus A_q\right)} \max \left( \|\mathbf{q}, \mathbf{e}^K\| - \|\mathbf{q}, \mathbf{e}'^K\| + \epsilon, 0 \right),
\end{equation}
where $e$ and $e'$ respectively denote an answer entity and a non-answer entity, $\|\mathbf{q}, \mathbf{e}^K\|$ denotes the Euclidean distance between the general question embedding $\mathbf{q}$ and the entity embedding $\mathbf{e}^K$, also $\epsilon \geq 0$ is the margin of the loss.

In inference, given a test question, we accordingly employ the GNN model to encode the KG and the question.
The top-\textit{N}\footnote{The value of \textit{N} is a hyperparameter that can be set according to the performance of the GNN on the training/validation set.} nearest neighbors of the question in the embedding space are retrieved as candidate answers and passed to the next step.

\textbf{Scalability Discussion.} The above general and layerwise question encoders are both implemented as lightweight GRU networks with the given question as input.
Their time complexity is not dependent on the KG.
As the length of questions in KGQA is insignificant compared with the size of the KG (e.g., the number of triples) and does not necessarily increase in larger datasets/KGs, the question encoders are highly scalable.
The attentional message propagation module mainly involves the computation of messages and attention scores for all triples in the KG.
Assuming the model configurations, including the number of layers and embedding dimensions, remain the same, its time complexity is $O(|T|)$ with respect to the size of the KG, where $|T|$ is the number of triples.
The gated embedding update module computes two attention scores and one updated embedding for each entity in the KG.
With respect to the KG, its time complexity is $O(|E|)$, where $|E|$ is the number of entities.
Therefore, the time complexities of the proposed modules increase linearly with respect to the number of triples and entities, respectively.
However, please note that it is not necessary to apply the GNN over the whole underlying KG.

When answering a given question, we only need to consider a relatively much smaller subgraph of the KG that covers the reasoning subgraph and all potential answer entities of the question.
Analogous to QA-GNN~\citep{DBLP:conf/naacl/YasunagaRBLL21} and GreaseLM~\citep{DBLP:journals/corr/abs-2201-08860}, this subgraph can be extracted by including triples that are few hops away from the topic entities.
This number of hops can be estimated based on the complexity of the question~\citep{DBLP:conf/esws/WangRCB24}.
Also, this retrieved subgraph can be further pruned by removing entities that are unlikely to be relevant to the question based on Personalized PageRank (PPR)~\citep{DBLP:conf/www/Haveliwala02}, as proposed in GraftNet~\citep{DBLP:conf/emnlp/SunDZMSC18}.
In this way, the GNN-based reasoning is practically scalable to very large KGs, as it only encodes the question-relevant part of the KG.

\subsection{Step-II: LM-based Explicit Reasoning}
\label{subsection:step_2}

\begin{figure*}[t]
    \centering
    \includegraphics[trim=5pt 0pt 0pt 0pt, clip, width=1\textwidth]{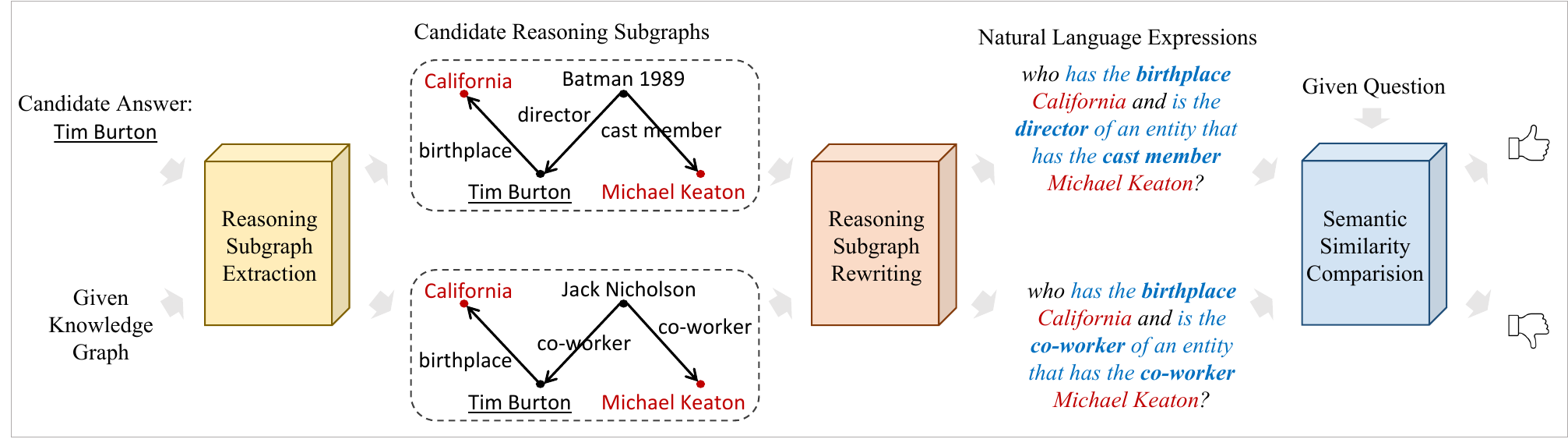}
    \caption{An overview of Step-II, assuming \texttt{Tim Burton} is extracted as a candidate answer regarding the example question and KG in \cref{fig:introduction}.} 
    \label{fig:step_2}
\end{figure*}

Based on the example question in \cref{fig:introduction}, an overview of this step is depicted in \cref{fig:step_2}.
We first extract subgraphs from the given KG that consist of paths linking \texttt{Tim Burton} (i.e., a candidate answer) to \texttt{California} and \texttt{Michael Keaton} (i.e., the topic entities).
Then, we rewrite the subgraphs into two natural language expressions and compare them with the given question to select the final reasoning subgraph according to their semantic similarity to the question.

\textbf{Reasoning subgraph extraction.}
Regarding each candidate answer, we utilize Graph-Tool\footnote{\url{https://graph-tool.skewed.de/}} to extract paths in the KG that link the candidate answer to any of the topic entities mentioned in the given question.
Then, we generate all candidate reasoning subgraphs based on the combinations of these paths.
Each candidate reasoning subgraph should contain one specific candidate answer that is connected to all topic entities.
Please note that we only consider the paths that are within a threshold of maximum length, which can be pre-defined according to the estimated complexity of the given question~\citep{DBLP:conf/esws/WangRCB24}.

\textbf{Reasoning subgraph rewriting.} 
In this module, we denote the set of candidate reasoning subgraphs as $G_q = \{\mathcal{G}_c\}$, where $\mathcal{G}_c$ denotes one specific subgraph.
We propose \cref{alg:subgraph_rewriting} to rewrite the subgraphs in $G_q$ into a set of natural language expressions, denoted as $\NL(G_q)$.
Specifically, in \cref{alg:subgraph_rewriting} - Line 3, the function $\WhPred(q)$ returns the wh-word used in the question $q$.
If there is no wh-word in $q$, this function always returns ``\textit{what}.''
In \cref{alg:subgraph_rewriting} - Line 4, the path $P$ refers to a path that is extracted in the above process and used to assemble the subgraph $\mathcal{G}_c$.
Also, the function $label(\cdot)$ returns the label of an input entity/relation, and the $+$ sign denotes the concatenation of strings.
Please note that the number of generated expressions is usually substantially smaller than the number of reasoning subgraphs, as the intermediate entities and candidate answers are not explicitly expressed.

\begin{algorithm}[t]
\caption{Reasoning subgraph rewriting.}
\label{alg:subgraph_rewriting}
\begin{algorithmic}[1]

\Require \Longunderstack[l]{Question $q$, candidate reasoning subgraph set $G_q$,\\ and the set $E_q$ including all topic entities.}
\Ensure Natural language expression set $\NL(G_q)$.

\State $\NL(G_q) \leftarrow$ an empty set
\For{subgraph $\mathcal{G}_c \in G_q$}
    \State $nl \leftarrow \WhPred(q)$
    \For{path $P$ in $\mathcal{G}_c$}
        \For{triple $(e_h, r, e_t)$ in $P$}
            \If{$r$ is an inversed relation}
                \State $nl \leftarrow nl + $ ``\textit{ is the }'' $ + \Label(r) + $ ``\textit{ of}''
            \Else
                \State $nl \leftarrow nl + $ ``\textit{ has the }'' $+ \Label(r)$
            \EndIf
            \If{$e_t \in E_q$}
                \State $nl \leftarrow nl + $ ``\textit{ }'' $ + \Label(e_t)$
            \Else
                \State $nl \leftarrow nl + $ ``\textit{ an entity that}''
            \EndIf
        \EndFor
        \State Remove $P$ from $\mathcal{G}_c$
        \If{$\mathcal{G}_c$ is not empty}
            \State $nl \leftarrow nl + $ ``\textit{ and}''
        \EndIf
    \EndFor
    \State Add $nl$ to $\NL(G_q)$
\EndFor

\end{algorithmic}
\end{algorithm}

\textbf{Semantic similarity comparison.}
In this step, we first use the Sentence Transformer~\citep{DBLP:conf/emnlp/ReimersG19} to encode the given question $q$ and natural language expressions in $\NL(G_q)$ into embedding vectors.
The output question embedding is denoted as $\mathbf{q}_t \in \mathbb{R}^{d_t}$, where $d_t$ is the output dimension of the Sentence Transformer.
For an expression $nl \in \NL(G_q)$, its output embedding is denoted as $\mathbf{nl} \in \mathbb{R}^{d_t}$.
We select the optimal expression $nl'$ that is most semantically similar to the given question according to the output embeddings:
\begin{equation}
    \label{equ:optimal_expression}
    nl' = \ArgMax_{nl \in \NL(G_q)} \Similarity\left(\mathbf{nl}, \mathbf{q}_t\right),
\end{equation}
where $\Similarity(\mathbf{nl}, \mathbf{q}_t)$ denotes the cosine similarity between $\mathbf{nl}$ and $\mathbf{q}_t$.
Finally, we return the subgraph corresponding to $nl'$ as the final reasoning subgraph.
Also, its contained answer entity would be the final answer.
In practice, there could be several subgraphs corresponding to $nl'$.
All these subgraphs and their contained answer entities could be considered as correct.
However, if a more restricted set of answers is desired, the distance of these answers to the given question in the embedding space of Step-I can be used for ranking.
It is worth mentioning that the length of the natural language expressions depends on the size of the candidate reasoning subgraphs.
Considering the efficiency and scalability of this step, we constrained this size by limiting the length of extracted paths in the above subgraph extraction step.

\begin{algorithm}
\caption{Prediction of positive and negative expressions.}
\label{alg:pos_neg_expression_prediction}
\begin{algorithmic}[1]

\Require \Longunderstack[l]{KG $\mathcal{G}$, a training/validation question $q$, ground-truth\\ answer set $A_q$, the set of reasoning subgraphs extracted\\ for both ground-truth and candidate answers: $G_q'$.}
\Ensure \Longunderstack[l]{Predicted positive set $\NL_p(q)$ and negative set \\$\NL_n(q)$ of natural language expressions.}

\State $\NL_p(q), \NL_n(q) \leftarrow $ empty sets

\State $nl2ans, nl2vote \leftarrow$ empty dictionary objects

\For{subgraph $\mathcal{G}_c \in G_q'$}
    \State $Q(\mathcal{G}_c) \leftarrow $ \Longunderstack[l]{convert $\mathcal{G}_c$ into a query by turning the \\included answer entity into a variable}

    \State $A(\mathcal{G}_c) \leftarrow $ \Longunderstack[l]{a set of entities that can be obtained by \\executing $Q(\mathcal{G}_c)$ over $\mathcal{G}$}
    
    \State $nl \leftarrow$ \Longunderstack[l]{the natural language expression generated \\based on $\mathcal{G}_c$ using \cref{alg:subgraph_rewriting}}

    \If{$nl \notin keys(nl2ans)$}
        \State $nl2ans[nl] \leftarrow $ an empty set
    \EndIf

    \For{entity $e' \in A(\mathcal{G}_c)$}
        \State Add $e'$ to $nl2ans[nl]$
    \EndFor
    \EndFor

\For{$nl \in $ $keys(nl2ans)$}
    \State $nl2vote[nl] \leftarrow \left| nl2ans[nl] \cap A_q \right| - \left| nl2ans[nl] \setminus A_q \right|$
\EndFor

\State $max\_vote \leftarrow $ \Longunderstack[l]{maximum vote of any $nl \in keys(nl2vote)$}
\State $min\_len \leftarrow $ \Longunderstack[l]{minimum number of entity/relation ment-\\ions in the expressions that have $max\_vote$}

\For{$nl \in $ $keys(nl2vote)$}
    \If{$nl2vote[nl] = max\_vote \land \Length(nl) = min\_len$}
        \State Add $nl$ to $\NL_p(q)$
    \Else
        \State Add $nl$ to $\NL_n(q)$
    \EndIf
\EndFor

\end{algorithmic}
\end{algorithm}

\textbf{Sentence Transformer Fine-tuning.}
The Sentence Transformer was pre-trained on the semantic search task~\citep{DBLP:conf/emnlp/ReimersG19} that is different from the above similarity comparison.
To mitigate this gap, we fine-tune the Sentence Transformer based on the available weak supervision (i.e., question-final answer pairs).

We propose \cref{alg:pos_neg_expression_prediction} to generate a set of relatively positive expressions $\NL_p(q)$ and a set of relatively negative expressions $\NL_n(q)$ for each training question $q$.
Specifically, we extract candidate reasoning subgraphs for the question using both the ground truth answers and the candidate answers obtained by performing the above GNN encoding.
Then, we convert each extracted subgraph into a query by turning its included answer entity into a variable.
By executing this query over the KG, we can get a set of entities---the entities that satisfy the semantic constraints of the extracted subgraph.
Therefore, the more these entities belong to the correct answers, the more positively this extracted subgraph would be considered.
Specifically, we consider entities being the correct answers as ``up-votes,'' while other entities being ``down-votes,'' as computed in \cref{alg:pos_neg_expression_prediction} - Line 15.
Finally, we add the most concise natural language expressions with the highest ``net-votes'' to the relatively positive set $\NL_p(q)$.
All the other natural language expressions are added to the relatively negative set $\NL_n(q)$.
The loss for fine-tuning the Sentence Transformer is defined as a triplet loss:
\begin{equation}
    \label{equ:fine_tuning_loss}
    l(q) = \sum_{\left(nl_p, nl_n\right) \in \NL'} \max \left( \Similarity\left(\mathbf{nl}_n, \mathbf{q}_t\right) - \Similarity\left(\mathbf{nl}_p, \mathbf{q}_t\right) + \epsilon', 0 \right),
\end{equation}
where $\NL' = \NL_p(q) \times \NL_n(q)$, $nl_p$ and $nl_n$ respectively denote a relatively positive expression and a relatively negative expression, $\Similarity(\cdot)$ denotes the cosine similarity function, and $\epsilon'$ is a non-negative margin.

Please note that our method is not limited to weakly supervised learning.
When the annotations of reasoning subgraphs are available, we can skip \cref{alg:pos_neg_expression_prediction} and directly use the natural language expressions of the ground-truth annotations as $\NL_p(q)$.
The expressions of other extracted candidate reasoning subgraphs can be used as $\NL_n(q)$.

\section{Experiments}

In this section, we conduct extensive experiments to evaluate GNN2R and compare it with recent methods.

\subsection{Effectiveness Evaluation}
\label{subsection:effectiveness}

\begin{table}[t]
\centering
\caption{The number of questions (in per split) and the average number of entities/relations/triples in question-specific subgraphs.}
\label{table:dataset_statistics}
\begin{tabular}{@{}ccccccc@{}}
\toprule
Dataset & \multicolumn{3}{c}{Question Set} & \multicolumn{3}{c}{KG Subgraph (Average)} \\ \cmidrule(l){2-7} 
        & \#Train   & \#Valid   & \#Test   & \#Entity    & \#Relation    & \#Triple   \\ \midrule
WQSP    & 2,848     & 250       & 1,639    &   1,459       &   296       &   4,347         \\
CWQ     & 27,639    & 3,519     & 3,531    &   1,473       &   292          &  4,751          \\
PQ-2hop     & 1,527    & 190     & 191    &   1,056       &   13          &  1,211  \\  
PQ-3hop     & 4,159    & 519     & 520    &   1,836       &   13          &  2,839  \\  
PQL-2hop     & 1,276    & 159     & 159    &   5,034       &   363          &  4,247 \\   
PQL-3hop     & 825    & 103     & 103    &   6,505       &   411          &  5,597 
\\
\bottomrule
\end{tabular}
\end{table}

We used four commonly-adopted KGQA benchmark datasets: WebQuestionsSP (WQSP)~\citep{DBLP:conf/acl/YihRMCS16}, ComplexWebQuestions (CWQ)~\citep{DBLP:conf/naacl/TalmorB18}, PathQuestion (PQ)~\citep{DBLP:conf/coling/ZhouHZ18}, and PathQuestion-Large (PQL)~\citep{DBLP:conf/coling/ZhouHZ18}.
WQSP and CWQ contain questions of varying complexity that require reasoning over up to two and four hops
, respectively.
PQ and PQL are split into PQ-2/3hop and PQL-2/3hop according to the required reasoning hops by the questions.
The underlying KG is Freebase~\citep{DBLP:conf/sigmod/BollackerEPST08}.
Based on prior work~\citep{DBLP:conf/acl/YihCHG15,DBLP:conf/www/Haveliwala02,DBLP:conf/wsdm/HeL0ZW21,DBLP:conf/emnlp/SunDZMSC18,DBLP:conf/coling/ZhouHZ18}, topic entities and a KG subgraph consisting of question-relevant triples can be extracted and are available for each benchmark question.
We report detailed statistics of the QA datasets in \cref{table:dataset_statistics}.

In our experiments, the number of GNN layers was always set to 3.
The margin in \cref{equ:gnn_loss} was set to 1.0 for WQSP, PQ-3hop, and both subsets of PQL, while 0.5 for CWQ and PQ-2hop.
In the first step of GNN2R, the number of candidate answers to select was 10 for WQSP, 25 for CWQ, 20 for PQ-2hop, and 5 for PQ-3hop and both subsets of PQL.
The margin in \cref{equ:fine_tuning_loss} was set to 0.1 for WQSP, CWQ, and PQ-3hop, 0.8 for PQ-2hop and PQL-2hop, and 0.05 for PQL-3hop.
These hyperparameters were determined via empirical grid search.

We compare with six baselines: EmbedKGQA~\citep{DBLP:conf/acl/SaxenaTT20}, GraftNet~\citep{DBLP:conf/emnlp/SunDZMSC18}, Relation Learning~\citep{DBLP:conf/emnlp/YanLWZDZWX21}, TransferNet~\citep{DBLP:conf/emnlp/ShiC0LZ21}, NSM~\citep{DBLP:conf/wsdm/HeL0ZW21}, and SR+NSM~\citep{DBLP:conf/acl/ZhangZY000C22}.
EmbedKGQA, GraftNet, and Relation Learning belong to the retrieval-based methods introduced in \cref{sec:related_work}.
The other three baselines, i.e., TransferNet, NSM, and SR+NSM, are reasoning-based methods that can provide their reasoning process as answer explanations.
In the weakly-supervised multi-hop KGQA task, TransferNet, NSM, and SR+NSM are the SOTA methods.

\begin{table*}[t]
{\centering
\caption{Overall performance (in terms of Hits@1 and F1 score) on benchmark datasets. (best performance in \textbf{bold}, second best \underline{underlined}, also percentage changes of our model with respect to the best baseline are shown at the bottom)
Results marked with \text{*} are the best-reported results known to us from existing publications. Other results are obtained from our experiments using released source code and, when available, pre-trained models from the authors. In cases where pre-trained models are not available, we adopt the best model and training configurations reported in original papers and source code.
}
\label{table:overal_performance}
\begin{tabular}{@{}ccccccc@{}}
\toprule
Method  &
\multicolumn{2}{c}{WQSP}  &
\multicolumn{2}{c}{CWQ}  &
\multicolumn{2}{c}{PQ-2hop} \\ \cmidrule(l){2-7}
 & Hits@1 & F1 & Hits@1 & F1 & Hits@1 & F1 \\
\midrule
EmbedKGQA & 66.6\textsuperscript{*} & 51.3 & 21.7 & 18.8 & 90.1 & 89.3 \\
GraftNet & 68.7\textsuperscript{*} & 62.3\textsuperscript{*} & 36.8\textsuperscript{*} & 
32.7\textsuperscript{*} & 90.6 & 89.3 \\
Relation Learning & 72.9\textsuperscript{*} & 64.5\textsuperscript{*} & 35.1 & 31.3 & \textbf{99.5} & \underline{98.4} \\
TransferNet & 71.4\textsuperscript{*} & 66.1 & 48.6\textsuperscript{*} & 36.0 & 91.1 & 91.1 \\
NSM & \underline{74.3}\textsuperscript{*} & \underline{67.4}\textsuperscript{*} & 48.8\textsuperscript{*} & 44.0\textsuperscript{*} & \underline{94.2} & 93.7 \\
SR+NSM & 69.5\textsuperscript{*} & 64.1\textsuperscript{*} & \underline{50.2}\textsuperscript{*} & \underline{47.1}\textsuperscript{*} & 93.7 & 93.5 \\ 
\midrule
GNN2R (Step-I) & 68.9 & 46.4 & 48.8 & 43.3 & 96.9 & 95.5 \\
GNN2R (Step-I+Step-II) & \textbf{75.8} & \textbf{67.6} & \textbf{53.8} & \textbf{51.7} & \textbf{99.5} & \textbf{99.5} \\
 & $\uparrow$ 2.0\% & $\uparrow$0.3\% & $\uparrow$ 7.2\% & $\uparrow$ 9.8\% &  0.0\% & $\uparrow$ 1.1\% \\
\bottomrule
\end{tabular}

\begin{tabular}{@{}ccccccc@{}}
\toprule
Method & \multicolumn{2}{c}{PQL-2hop} & 
\multicolumn{2}{c}{PQ-3hop} & \multicolumn{2}{c}{PQL-3hop} \\ \cmidrule(l){2-7}
 & Hits@1 & F1 & Hits@1 & F1 & Hits@1 & F1 \\
\midrule
EmbedKGQA & 79.2 & 76.5 & 86.2 & 84.5 & 61.2 & 58.9 \\
GraftNet & 87.1 & 74.3 & 93.1 & 92.7 & 84.0 & 65.1 \\
Relation Learning & \underline{98.1} & \underline{97.7} & 59.0 & 53.9 & \underline{85.4} & \underline{85.7} \\
TransferNet & 54.7 & 50.3 & 96.5 & \underline{96.7} & 62.1 & 64.3 \\
NSM & 74.2 & 75.0 & \underline{97.1} & 96.2 & 67.0 & 68.3 \\
SR+NSM & 76.7 & 75.7 & 96.7 & \underline{96.7} & 68.0 & 66.3 \\ 
\midrule
GNN2R (Step-I) & 99.4 & 80.9 & 94.8 & 88.8 & 91.3 & 84.5 \\
GNN2R (Step-I+Step-II) & \textbf{99.4} & \textbf{98.4} & \textbf{99.2} & \textbf{99.0} & \textbf{97.1} & \textbf{96.9} \\
 & $\uparrow$ 1.3\% & $\uparrow$0.7\% & $\uparrow$ 2.2\% & $\uparrow$ 2.4\% & $\uparrow$ 13.7\% & $\uparrow$ 13.1\% \\
\bottomrule
\end{tabular}
}

\end{table*}

The overall performance of our model is reported in \cref{table:overal_performance}.
Following the baselines, we use Hits@1 and F1 score (both in percentage form) as evaluation metrics.
Hits@1 refers to the proportion of test questions for which the returned \textit{top-1} answers are correct.
Since there could be multiple correct answers for a given question, the F1 score is calculated based on the precision and recall of all returned answers.
In the row \textbf{GNN2R (Step-I)}, we report the results with only the first step of our model.
Specifically, for a test question, we return its nearest-neighboring entity in the computed embedding space as the \textit{top-1} answer.
For calculating the F1 score, other entities within a threshold of distance\footnote{This distance is calculated by multiplying the shortest distance by a pre-defined threshold.} are returned as all possible answers.
In the row \textbf{GNN2R (Step-I + Step-II)}, we report the results of the complete model.
Final answers (entities) can be obtained according to the retrieved reasoning subgraphs.
Similarly, the entity that has the shortest distance in Step-I is returned as the \textit{top-1} answer, while other entities within a threshold of distance are considered as the final answers.

The following can be observed from \cref{table:overal_performance}:

\noindent 1) GNN2R achieves superior performance on all benchmark datasets in comparison with all six baselines.
Especially on CWQ and PQL-3hop, which are the two most challenging datasets in our experiments regarding question complexity and the size of KGs, significant improvements (7.2\%/9.8\% on CWQ and 13.7\%/13.1\% on PQL-3hop with respect to Hits@1/F1) have been achieved in comparison to previous best-performing baselines.
Also, PQ and PQL are nearly solved by GNN2R.
To the best of our knowledge, we are the first and, thus far, the only one to achieve this.\footnote{A public leaderboard is available at \url{https://github.com/KGQA/leaderboard/blob/gh-pages/freebase/path_question.md}.}
   
\noindent 2) GNN2R remains highly competitive even with only the first step (Step-I).
For example, on CWQ, Step-I achieves a similar performance as the previous SOTA method NSM and outperforms four other baselines.
On PQL-2hop and PQL-3hop, Step-I is ranked first among all baselines regarding Hits@1.
Also, if considering F1, Step-I is ranked second on PQ-2hop, PQL-2hop, and PQL-3hop.

\noindent 3) The performance of GNN2R can be substantially improved with the second step (Step-II).
Specifically, on the six benchmark datasets, average improvements of 5.7\% and 19.5\% regarding Hits@1 and F1 are achieved when Step-II is performed based on Step-I.
These results demonstrate the effectiveness of Step-II and the weakly-supervised fine-tuning method.
Moreover, it is noteworthy that the improvement in F1 is more substantial than that in Hits@1, which demonstrates that Step-II can effectively filter out incorrect candidate answers retrieved from Step-I.

\begin{table}[t]
\centering
\caption{Comparison of the average run-time (ms) for answering a question. (mean $\mu$ and standard deviation $\sigma$ of five runs)}
\label{table:efficiency_compare}
\begin{tabular}{ccccc}
\hline
         & GNN2R & GNN2R-fast & NSM   & TransferNet \\ \hline
$\mu$    & 222.1 & 200.9      & 241.4 & 223.5       \\
$\sigma$ & 1.3   & 0.9        & 4.7   & 18.4        \\ \hline
\end{tabular}
\end{table}

\begin{table}[t]
\centering
\caption{The average run-time (ms) of all major phases of GNN2R in the process of answering a question. (mean $\mu$ and standard deviation $\sigma$ of five runs)
}
\label{table:efficiency}
\begin{tabular}{@{}llcc@{}}
\toprule
Step& Phases                                        & $\mu$   & $\sigma$ \\ \midrule
\textit{Step-I: GNN-based}   &       
Question \& KG Preprocessing                  & 22.1  & 0.2    \\
\textit{Coarse Reasoning} & GNN Encoding                                  & 5.7   & 0.0    \\
& Candidate Answer Selection                    & 3.4   & 0.1    \\ \midrule
\textit{Step-II: LM-based}
& Reasoning Subgraph Extraction                 & 160.1 & 1.2    \\
\textit{Explicit Reasoning}& Reasoning Subgraph Rewriting                  & 0.3   & 0.0    \\
& Semantic Similarity Comparison                & 28.7  & 0.1    \\ \midrule
Total & (\textit{including data loading, etc.})  & 222.1 & 1.3    \\ \bottomrule
\end{tabular}
\end{table}

\subsection{Efficiency Evaluation}
\label{subsection:efficiency}

In this section, we aim to evaluate the efficiency of GNN2R in comparison with NSM and TransferNet---the SOTA methods without the retrieval of reasoning subgraphs.\footnote{SR+NSM aims to provide enhanced input subgraphs for NSM. It shares the same reasoning process as NSM and, therefore, is not particularly evaluated.}
As WQSP is collected from Google Suggest API~\citep{DBLP:conf/emnlp/BerantCFL13} with human evaluations on Amazon Mechanical Turk,\footnote{\url{https://www.mturk.com/}} we consider it a good representation of real-world questions and use it in this evaluation.
\cref{table:efficiency_compare} reports the average run-time of GNN2R, GNN2R-fast, NSM, and TransferNet on the test set of WQSP.
On average, GNN2R can answer a test question within 222.1ms.
\textbf{GNN2R-fast} refers to a faster version of GNN2R, which only considers the shortest paths between topic entities and candidate answers when extracting paths for generating candidate reasoning subgraphs in Step-II.
The Hits@1 of GNN2R-fast on WQSP slightly drops to 75.4, which is still higher than NSM, TransferNet, and SR+NSM.
It can be observed that GNN2R is substantially more efficient than NSM (8.0\% time reduction) and slightly more efficient than TransferNet.
GNN2R-fast is substantially more efficient than both NSM and TransferNet, with 16.8\% and 10.1\% time reductions, respectively.
Also, the run-times for GNN2R and GNN2R-fast seem to be more stable across different runs, as suggested by the smaller standard deviations.

In addition, we report the time costs associated with all major phases of GNN2R in \cref{table:efficiency}.
We can observe that both the GNN encoding and the LM-based semantic similarity comparison are highly efficient.
The most time-consuming process is the graph algorithm-based reasoning subgraph extraction.
Please note that the above evaluations were conducted on a Linux server with two Intel Xeon Gold 6230 CPUs (2.10GHz) and one NVIDIA GeForce RTX 2080 Ti GPU being used.
For NSM and TransferNet, we used the code released by the original authors.\footnote{\url{https://github.com/RichardHGL/WSDM2021_NSM} and \url{https://github.com/shijx12/TransferNet}}

\subsection{Ablation Study and Practical Implications}
\label{subsection:ablation_study}

\begin{table*}[t]
\centering
\caption{The performance of GNN2R in different degraded scenarios. (Hits@1 \%)}
\label{table:ablation_results}
\begin{tabular}{@{}cp{0.48\linewidth}cc@{}}
\toprule
 & Degraded Scenarios & WQSP & CWQ \\ \midrule
 \multirow{4}{*}{Step-I} & Original Step-I & 68.9 & 48.8 \\
 & Step-I without Gated Embedding Update Module (w/o Gated Embedding) & 62.8 ($\downarrow$ 8.9\%) & 47.5 ($\downarrow$ 2.7\%) \\
 & Step-I without Attentional Message Propagation (w/o Attentional Message) & 60.8 ($\downarrow$ 11.8\%) & 39.7 ($\downarrow$ 18.6\%) \\
 & Step-I w/o both Gated Embedding and Attentional Message & 56.4 ($\downarrow$ 18.1\%) & 38.6 ($\downarrow$ 20.9\%) \\ \midrule
\multirow{5}{*}{Step-II} & Original Step-II + Original Step-I & 75.8 & 53.8 \\
 & Step-II without Fine-tuning + Original Step-I & 60.1 ($\downarrow$ 20.7\%) & 39.1 ($\downarrow$ 27.3\%) \\
 & Original Step-II + Step-I w/o Gated Embedding & 73.1 ($\downarrow$ 3.6\%) & 53.3 ($\downarrow$ 0.9\%) \\
 & Original Step-II + Step-I w/o Attentional Message & 73.1 ($\downarrow$ 3.6\%) & 52.4 ($\downarrow$ 2.6\%) \\
 & Original Step-II + Step-I w/o both Gated Embedding and Attentional Message & 70.7 ($\downarrow$ 6.7\%) & 52.4 ($\downarrow$ 2.6\%) \\ \bottomrule
\end{tabular}

\end{table*}

We attribute the superior performance of GNN2R to the following aspects of the model and empirically examine their contributions in corresponding degraded scenarios:\\
1) In Step-I, the gated embedding update module (cf. the lower-side blue box in \cref{fig:step_1}) controls how much the messages received by an entity should be added to the entity's embedding with reference to the general question embedding.
This ensures that the embeddings of answer entities keep question-relevant messages while not being ``diluted'' by irrelevant messages across multiple layers of encoding.
To investigate the contribution of this aspect, we remove the gated embedding update module from the GNN and directly update entity embeddings with their received messages, which is a common setup of existing GNNs~\citep{DBLP:conf/iclr/VashishthSNT20,DBLP:conf/esws/SchlichtkrullKB18,DBLP:journals/corr/abs-1710-10903}.
We isolate the GNN from the two-step reasoning process and examine its performance on WQSP and CWQ.
The Hits@1 results of the original GNN and this degraded version are reported in the first and second rows of \cref{table:ablation_results} - Step-I block, respectively.
In comparison with the original GNN, we observe performance drops of 8.9\% and 2.7\% on WQSP and CWQ, respectively.
\\
2) For complex multi-hop questions, the GNN may need to focus on different semantic aspects of the question at different encoding layers. To address this, we propose the layerwise question encoder (illustrated by the yellow box in \cref{fig:step_1}), which captures potentially distinct semantic aspects of input questions and represents them as layerwise reference embeddings. Subsequently, the stepwise attentional message propagation module (the upper-side blue box in \cref{fig:step_1}) leverages these embeddings to pass question-relevant messages toward the answer entities.
In the third row of the Step-I block in \cref{table:ablation_results}, we report the performance of the GNN after removing the layerwise question encoding and the corresponding attentional message propagation. 
The performance declines significantly by 11.8\% and 18.6\% on WQSP and CWQ, respectively, demonstrating the effectiveness of the layerwise encoder in our GNN.
Furthermore, the last row in the Step-I block of \cref{table:ablation_results} shows results for the GNN without both gated embedding updates and attentional message propagation, reducing the model to a conventional GNN structure~\citep{DBLP:conf/iclr/VashishthSNT20,DBLP:conf/esws/SchlichtkrullKB18,DBLP:conf/iclr/KipfW17}.
In this scenario, the performance drops significantly---by 18.1\% and 20.9\% on WQSP and CWQ, respectively.
\\
3) Another major contributing factor is that the proposed weakly-supervised fine-tuning approach can effectively mitigate the gap between the pre-trained LM and the semantic comparison task in Step-II.
In the second row of \cref{table:ablation_results} - Step-II block, we freeze the original Step-I and examine the performance of GNN2R when the LM is not fine-tuned.
Substantial performance drops can be observed on both datasets, i.e., -20.7\% on WQSP and -27.3\% on CWQ, highlighting the contribution of the weakly-supervised fine-tuning.
\\
4) The last three rows in \cref{table:ablation_results} present GNN2R's final QA performance with the three types of degraded GNN in Step-I: Step-I w/o Gated Embedding, Step-I w/o Attentional Message, and Step-I w/o both Gated Embedding and Attentional Message.
In comparison with the original GNN2R model, evident performance drops can be observed, demonstrating the positive impact of the proposed gated embedding update module and the attentional message propagation on the overall QA task.
In addition, comparing these three rows with the last three rows in \cref{table:ablation_results} - Step-I block, we can observe substantial improvement, which demonstrates that the proposed Step-II can effectively improve the final QA performance even when the results from Step-I are degraded.
\\
Based on these results, we summarize the following practical implications:
First, when using GNN to encode KGs regarding a given complex question, it is important to use the question as a reference in the GNN's message propagation, which helps the GNN focus on question-relevant information.
Second, such question reference can be obtained by layer-wise question encoders that learn to extract distinct semantic aspects of the given question in different GNN layers, contributing to the final entity embedding-based retrieval of correct answers.
Third, to avoid accumulating messages that are irrelevant to the given question in the answer entities' embeddings, we can use a general question encoder to provide general reference information for updating entity embeddings in the GNN encoding.
Lastly, additional ranking and filtering of the coarse results from GNN encoding, such as our proposed Step-II, can substantially improve the model's performance in the overall QA task.

\subsection{Evaluation of Generated Reasoning Subgraphs}
\label{sec:explainability_eval}

In this section, we present a qualitative case study and a quantitative evaluation based on the above benchmark datasets to demonstrate the quality of generated reasoning subgraphs.

\begin{figure}[t]
    \centering
    \includegraphics[trim=5pt 8pt 0pt 0pt, clip, width=0.9\textwidth]{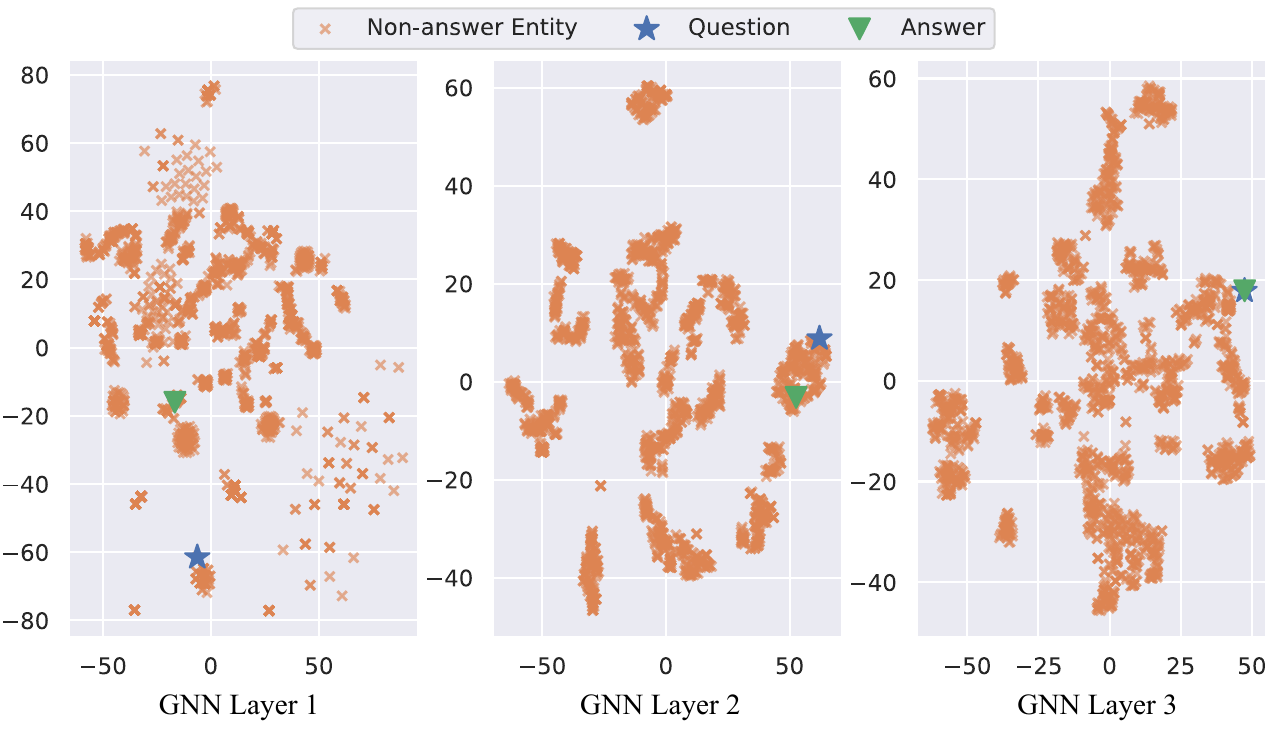}
    \caption{The visualization of the embedding space after each of the three GNN layers for answering the question ``\textit{what is the name of the place of birth of offspring of Henriette Adelaide of Savoy's husband?}'' The non-answer entities in the KG, the general question embedding, and the answer entity are represented as orange crosses, blue stars, and green triangles, respectively.}
    \label{fig:embed_space}
\end{figure}

\textbf{Positive Cases.}
We present a positive case study based on the following two complex questions:\\
Q1: ``\textit{What is the name of the place of birth of offspring of Henriette Adelaide of Savoy's husband?}''\\
Q2: ``\textit{Name the Super Bowl in which the New York Giants defeated the Buffalo Bills for the championship?}''\\
In the GNN encoding of Step-I, we expect to obtain a joint embedding space where answer entities are close to the question.
The intuition is to update the embeddings of answer entities with the messages propagated from the question-relevant triples.
For example, regarding Q1, we aim to update the embedding of the answer \texttt{Munich} with messages from the three triples: (\texttt{Henriette Adelaide of Savoy}, \texttt{\textbf{spouse}}, \texttt{Ferdinand Maria Elector of Bavaria}), (\texttt{Ferdinand Maria Elector of Bavaria}, \texttt{\textbf{children}}, \texttt{Joseph Clemens of Bavaria}), and (\texttt{Joseph Clemens of Bavaria}, \texttt{\textbf{place\_of\_birth}}, \texttt{Munich}).
In \cref{fig:embed_space}, we employ t-SNE~\citep{maaten2008visualizing} to visualize the embedding space of each of the three GNN layers for answering Q1.
The non-answer entities in the KG, the general question embedding, and the answer entity \texttt{Munich} are represented as orange crosses, blue stars, and green triangles, respectively.
It can be observed that the answer entity gets closer to the question after each layer of encoding and finally becomes the nearest neighbor of the question after three layers.
This is in line with our intuition that messages from the above three triples are hop-by-hop passed to the answer entity, demonstrating the reasoning process of GNN2R Step-I in the embedding space.

For Q2, the candidate answers computed from Step-I include \texttt{Super Bowl XXV}, \texttt{Super Bowl XLVI}, \texttt{Super Bowl XXI}, \texttt{Super Bowl XLII}, and \texttt{2008 NFC Championship Game}.
In Step-II, GNN2R generates candidate reasoning subgraphs for each candidate answer by extracting paths that link the candidate answer with the topic entities: \texttt{Super bowl}, \texttt{Buffalo Bills}, and \texttt{New York Giants}.
There are 118 candidate reasoning subgraphs in total.
The one with the highest semantic similarity with regard to the given question is: \{(\texttt{Super Bowl XXV}, \texttt{\textbf{notable\_types}}, \texttt{Super bowl}), (\texttt{Super Bowl XXV}, \texttt{\textbf{runner\_up}}, \texttt{Buffalo Bills}), (\texttt{Super Bowl XXV},\\ \texttt{\textbf{champion}}, \texttt{New York Giants})\}.
This subgraph does not directly represent the relationship between \texttt{New York Giants} and \texttt{Buffalo Bills} as expressed by the question: ``\textit{the New York Giants defeated the Buffalo Bills}.''
Instead, it implies this information by noting \texttt{New York Giants} as the champion while \texttt{Buffalo Bills} as the runner-up.
This case demonstrates that our method can extract potentially correct reasoning subgraphs based on the existing KG structures and also effectively examine the semantic similarity of the subgraphs based on the fine-tuned LM.

\textbf{Negative Cases.}
In the following, we summarize typical cases where the method tends to fail:\\
1) Questions that involve processing entities' attributes, e.g., counting, sorting, filtering, and comparing.
Examples include ``\textit{out of the countries with territories in Oceania, which has the smallest number of people in their army,}'' ``\textit{which country in the Caribbean has the smallest ISO number,}'' and ``\textit{which country that is situated in the ASEAN Common Time Zone has the largest population.}''
As a retrieval-based method, GNN2R indeed lacks the capability of performing these operations, which we also report in \cref{sec:limitations}.\\
2) Questions that require the entities' class information.
For example, the reasoning subgraph generated by GNN2R for the question ``\textit{what highschool did harper lee go to}'' is:
\{(\texttt{Huntingdon College}, \texttt{students\_graduates}, \texttt{m.0lwxmy1}\footnote{\texttt{m.0lwxmy1} is a compound value type entity.}), (\texttt{m.0lwxmy1}, \texttt{education\textsuperscript{-1}}, \texttt{Harper Lee})\}.
This subgraph is incorrect, as it ignores that the class of \texttt{Huntingdon College} is not a ``\textit{highschool}.''\\
3) Short questions with limited constraints. 
One example is ``\textit{what did st augustine do?}''
The candidate answers from Step-I indeed include many relevant entities, such as \texttt{The Confessions of St. Augustine} and \texttt{The Enchiridion Manual}.
However, the answers \texttt{Writer} and \texttt{Physician} are not returned.
This is due to the limited information in the question that GNN2R can use to align the answers with the question in the GNN encoding.

\begin{table*}[t]
\centering
\caption{Evaluation of explanations generated by GNN2R and other reasoning-based baselines.}
\label{table:subgraph_matching}
\begin{tabular}{@{}ccccccc@{}}
\toprule 
 & \multicolumn{3}{c}{PQ-2hop} & \multicolumn{3}{c}{PQL-2hop} \\ \cmidrule(l){2-7} 
 & Precision & Recall & F1 & Precision & Recall & F1 \\ \midrule
NSM & 0.14 & 0.28 & 0.18 & 0.10 & 0.16 & 0.12 \\
SR+NSM & 0.13 & 0.27 & 0.18 & 0.08 & 0.11 & 0.09 \\
TransferNet & \underline{0.89} & \underline{0.89} & \underline{0.89} & \underline{0.39} & \underline{0.45} & \underline{0.42} \\ \midrule
GNN2R & \textbf{0.97} & \textbf{0.97} & \textbf{0.97} & \textbf{0.84} & \textbf{0.69} & \textbf{0.76} \\
\multicolumn{1}{l}{} & \multicolumn{1}{l}{$\uparrow$ 9.0\%} & \multicolumn{1}{l}{$\uparrow$ 9.0\%} & \multicolumn{1}{l}{$\uparrow$ 9.0\%} & \multicolumn{1}{l}{$\uparrow$ 115.4\%} & \multicolumn{1}{l}{$\uparrow$ 53.3\%} & \multicolumn{1}{l}{$\uparrow$ 81.0\%} \\ \bottomrule
\end{tabular}

\begin{tabular}{@{}ccccccc@{}}
\toprule 
 & \multicolumn{3}{c}{PQ-3hop} & \multicolumn{3}{c}{PQL-3hop} \\ \cmidrule(l){2-7} 
 & Precision & Recall & F1 & Precision & Recall & F1 \\ \midrule
NSM & 0.13 & 0.23 & 0.17 & 0.13 & 0.14 & 0.14 \\
SR+NSM & 0.15 & 0.25 & 0.19 & 0.12 & 0.13 & 0.12 \\
TransferNet & \underline{0.84} & \textbf{0.94} & \underline{0.89} & \underline{0.50} & \underline{0.60} & \underline{0.55} \\ \midrule
GNN2R & \textbf{0.91} & \underline{0.89} & \textbf{0.90} & \textbf{0.89} & \textbf{0.63} & \textbf{0.74} \\
   & \multicolumn{1}{l}{$\uparrow$ 8.3\%} & \multicolumn{1}{l}{$\downarrow$ 5.3\%} & \multicolumn{1}{l}{$\uparrow$ 1.1\%} & \multicolumn{1}{l}{$\uparrow$ 78.0\%} & \multicolumn{1}{l}{$\uparrow$ 5.0\%} & \multicolumn{1}{l}{$\uparrow$ 34.5\%} \\ \bottomrule
\end{tabular}
\end{table*}

\textbf{Comparison with the SOTA Methods} We compare our reasoning subgraphs with explanations that the SOTA reasoning-based methods (i.e., NSM, SR+NSM, and TransferNet) can provide.

NSM and SR+NSM compute step-wise entity distributions during reasoning over KGs.
Entities with high probabilities during intermediate/final steps are assumed to be intermediate/answer entities in reasoning chains, which can be used as explanations.
The main drawback is that NSM and SR+NSM do not examine if the reasoning chains are semantically consistent with the given questions.
For example, for the question from WQSP ``\textit{who is gimli's father in the hobbit},'' the NSM model released by the authors provides the reasoning chain:
\{(\texttt{Gloin}, \texttt{species}, \texttt{Dwarf}), (\texttt{Dwarf}, \texttt{characters of this species}, \texttt{Gimli})\}, which indeed leads to the correct answer \texttt{Gloin} but is wrong by itself.
In contrast, this issue is alleviated in GNN2R, as we compare the semantics of reasoning subgraphs and given questions.
GNN2R can provide the correct explanation: \{(\texttt{Gloin}, \texttt{parents\textsuperscript{-1}}, \texttt{Gimli})\}.

TransferNet aims to pass scores from topic entities to final answers via KG triples.
The authors propose to extract relations with transfer scores higher than a threshold as the reasoning process.
However, this threshold is difficult to specify, as there are usually too many relations with close scores.
For example, for the question ``\textit{what state is saint louis university in},'' using the threshold 0.9 specified in the released code, TransferNet selects 204 + 225 relations in a two-step reasoning.
In the first step, the selected relations (scores) are \texttt{headquarters} (0.999), \texttt{institution} (0.997), \texttt{location} (0.997), and \texttt{citytown} (0.997).
In the second step, the top selected relations include \texttt{location} (1.000), \texttt{institution} (1.000), and \texttt{spouse\_s\_reverse} (1.000).
It is difficult for users to examine these relations.
In comparison, our reasoning subgraphs are specific and concise.

Furthermore, we use ground-truth annotations of reasoning subgraphs in PQ and PQL to examine the quality of generated reasoning subgraphs.
A comparison with NSM, SR+NSM, and TransferNet is reported in \cref{table:subgraph_matching}.
Specifically, we check if the triples within our reasoning subgraphs or the explanations of the baselines are present in the ground-truth annotations and calculate precision, recall, and F1 on the triple level.
We can observe that the quality of our generated reasoning subgraphs is substantially higher than that of the SOTA methods, except for the recall on PQ-3hop.
In comparison to the best-performing baseline, i.e., TransferNet, average improvements of 52.7\% (precision), 15.5\% (recall), and 31.4\% (F1) are achieved on the four datasets.
Also, on the datasets with less training data and bigger KGs, i.e., PQL-2hop and PQL-3hop, the improvements of GNN2R are more substantial.

\subsection{Robustness Evaluation}

In this section, we assess the robustness of GNN2R in relation to the quantity and quality of the training data.
Specifically, two series of experiments are conducted on PQL-2hop, PQL-3hop, and WQSP:
1) For each dataset, we randomly remove 25\%, 50\%, 75\%, and 95\% of the training questions.
Then, we train GNN2R with the reduced training set and evaluate the model on the original test set.
For PQL-2hop, PQL-3hop, and WQSP, the numbers of test questions are 159, 103, and 1,639, respectively, while the numbers of kept training questions are [957, 638, 319, 63], [618, 412, 206, 41], and [2031, 1354, 677, 135] in the four scenarios.
2) We randomly shuffle the words in the removed 25\%, 50\%, 75\%, and 95\% training questions and add these questions back to the training set.
GNN2R is then trained with the complete yet noisy training set and, again, evaluated on the original test set.

\begin{figure}[t]
    \centering
    \includegraphics[trim=8pt 0pt 0pt 0pt, clip, width=0.9\textwidth]{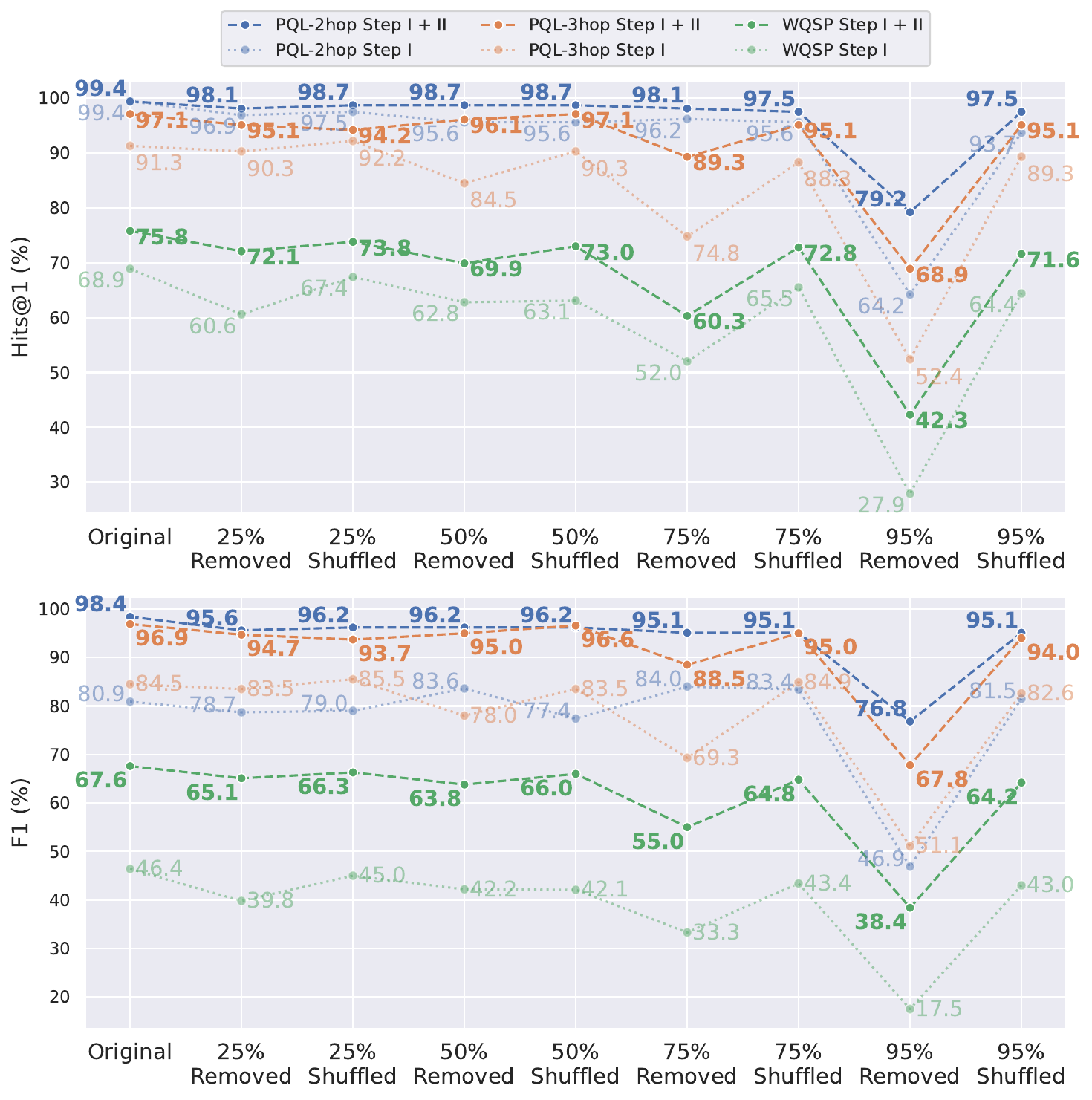}
    \caption{The performance of GNN2R (Hits@1 and F1 in percentage form) on PQL-2hop, PQL-3hop, and WQSP when 25\%, 50\%, 75\%, and 95\% training questions are directly removed or randomly shuffled.}
    \label{fig:data_removed_shuffled_plot}
\end{figure}

We plot the general performance of the trained models in \cref{fig:data_removed_shuffled_plot}, where the x-axes denote the different training scenarios, such as ``Original,'' ``25\% Removed,'' and ``25\% Shuffled.''
Hits@1 and F1 results are reported in the upper and lower subplots, respectively.
We use three different colors, i.e., blue, orange, and green, to depict the results of the three benchmarking datasets, i.e., PQL-2hop, PQL-3hop, and WQSP, respectively.
Also, the performance of the complete model (Step-I + Step-II) is plotted in bold lines, while the performance of only Step-I is in faded lines.
The following can be observed from \cref{fig:data_removed_shuffled_plot}:\\
1) The general performance of GNN2R indeed degrades when the training data becomes limited or noisy.
However, GNN2R demonstrated strong robustness.
According to the bold lines in \cref{fig:data_removed_shuffled_plot}, GNN2R maintains a relatively stable performance when up to 50\% of the training questions are removed or shuffled.
We only start observing more evident drops when 75\% or 95\% of the training questions are removed or shuffled.
However, these drops are still moderate compared with the drastic removal or shuffling of the training questions.
In the case of ``75\% Removed,'' the Hits@1/F1 drops on PQL-2hop, PQL-3hop, and WQSP are only -1.3\%/-3.4\%, -8.0\%/-8.7\%, and -20.5\%/-18.6\%, respectively.
Even in the case of ``95\% Removed,'' the Hits@1/F1 drops on the three datasets are limited to -20.3\%/-22.0\%, -29.0\%/-30.0\%, and -44.2\%/-43.2\%, respectively.
\\
2) By comparing the faded lines with the bold lines in the same color, we can observe that Step-I has similar performance changes as the complete model in different scenarios with different datasets, demonstrating the robustness of the GNN model.
Also, as the bold lines are higher than the corresponding faded lines in all degraded cases, GNN2R constantly achieves a performance improvement by performing Step-II after Step-I.
This shows the effectiveness of the proposed two-step strategy in scenarios with limited or noisy data.\\
3) The noise in the training data has a limited impact on the performance of GNN2R.
Even when 95\% of the training questions are randomly shuffled, the Hits@1/F1 of GNN2R only drops -1.9\%/-3.4\%, -2.1\%/-3.0\%, and -5.5\%/-5.0\% on PQL-2hop, PQL-3hop, and WQSP, respectively.
This demonstrates that the question encoders in Step-I and the LM in Step-II are able to capture and represent the correct semantics of given questions even when the order of the words is random.

\begin{figure}[t]
    \centering
    \includegraphics[trim=8pt 0pt 0pt 0pt, width=0.9\textwidth]{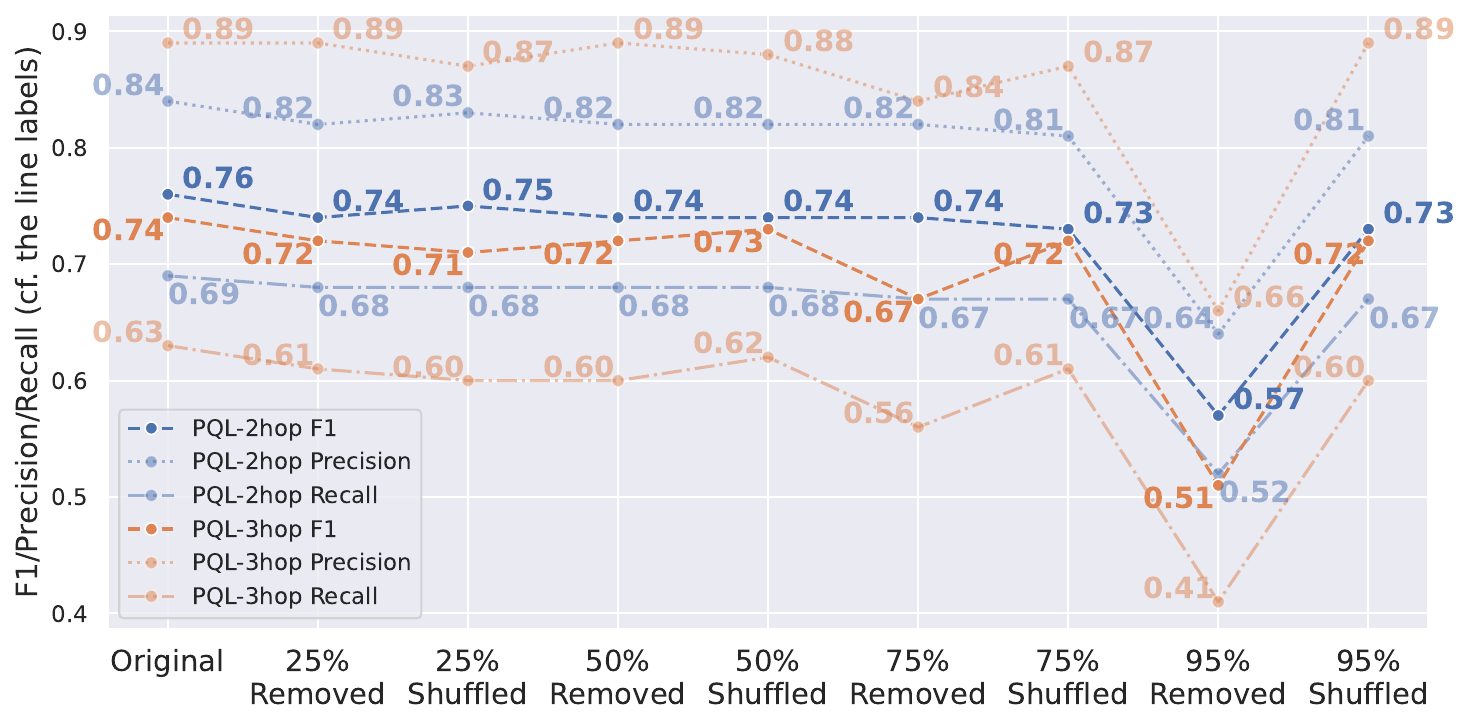}
    \caption{The quality of reasoning subgraphs generated by GNN2R (F1, Precision, and Recall) on PQL-2hop and PQL-3hop when 25\%, 50\%, 75\%, and 95\% training questions are directly removed or randomly shuffled.}
    \label{fig:subgraph_removed_shuffled_plot}
\end{figure}

In the following, we particularly examine the generated reasoning subgraphs in the above degraded cases using the ground-truth annotations of PQL-2/3hop.
Specifically, we check if the triples in the generated reasoning subgraphs are included in the ground-truth annotations and plot the precision, recall, and F1 of the triples in \cref{fig:subgraph_removed_shuffled_plot}.
We use the same colors as \cref{fig:data_removed_shuffled_plot} to represent the results on PQL-2/3hop.
Also, F1 is depicted in bold lines, while precision and recall are depicted in faded lines with different line styles.

According to \cref{fig:subgraph_removed_shuffled_plot}, similar to the above results, the quality of generated reasoning subgraphs remains robust when up to 50\% of the training questions are removed or shuffled: the performance drops are within -4.1\% on both datasets regarding F1.
Even with ``95\% Removed,'' GNN2R is still highly competitive compared with the best-performing baseline TransferNet regarding F1, i.e., 0.57/0.51 versus 0.42/0.55 (cf. \cref{table:subgraph_matching}) on PQL-2/3hop.
Also, by comparing the faded lines with the bold lines in the same color, we can observe that the performance regarding precision or recall is also robust.
Furthermore, similar to the above general results in \cref{fig:data_removed_shuffled_plot}, the noise in training has a limited impact on the quality of generated reasoning subgraphs.

The major reason for the above robust performance in \cref{fig:data_removed_shuffled_plot,fig:subgraph_removed_shuffled_plot} is the rich semantic information of the KG that is represented by entity/relation labels and graph structures.
The QA model does not need to memorize the knowledge required for answering given questions.
It only needs to learn how to reason out the answer based on the existing semantic knowledge, which is feasible to achieve with limited or noisy training data.
The GNN model in Step-I is adequately designed for this task.
It utilizes layerwise reference information from the given question to accordingly perform message propagation based on the label embeddings and graph structures of the KG.
Also, the LM in Step-II is effectively fine-tuned to utilize its pre-trained knowledge and language understanding capability to compare the semantic similarity between questions and reasoning subgraphs.

\begin{figure}[t]
    \centering
    \includegraphics[trim=25pt 25pt 0pt 0pt, clip, width=0.8\textwidth]{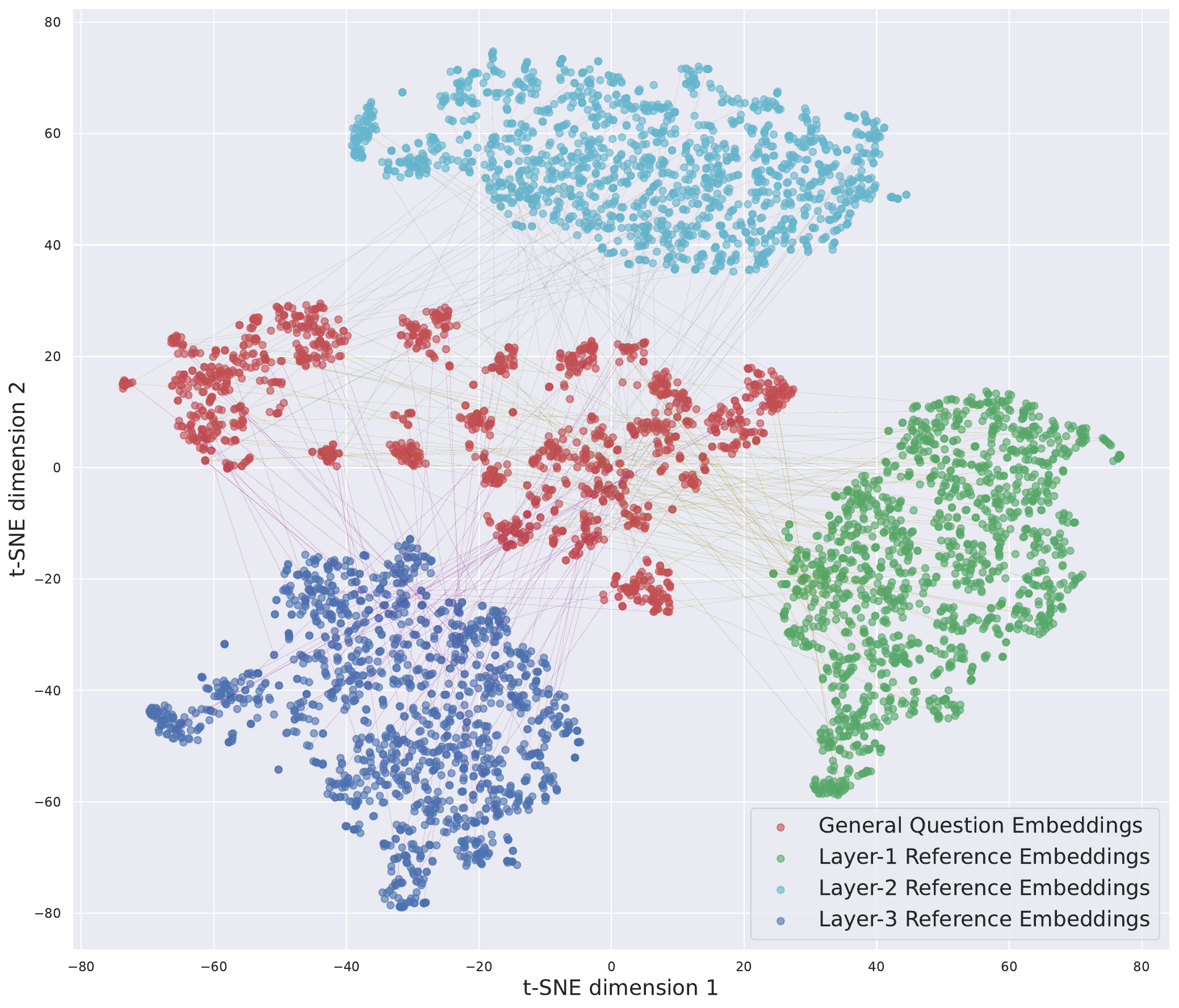}
    \caption{T-SNE visualization of the layerwise reference embeddings and general question embeddings of WQSP test questions. The layerwise reference embeddings of randomly selected 5\% questions are linked with their corresponding general question embeddings via colored lines.}
    \label{fig:reference_embed}
\end{figure}

\subsection{Visualization of Reference Information}
As mentioned in \cref{subsection:step_1}, the reference information computed in different GNN layers may have different foci and convey distinct aspects of the given question.
We first demonstrate this by visualizing the layerwise reference embeddings and the general question embeddings computed for the test questions in WQSP.
Specifically, we use t-SNE~\citep{maaten2008visualizing} to reduce the original dimensionality and plot the embeddings in \cref{fig:reference_embed}, where the general question embeddings and the reference embeddings of the first, second, and third GNN layers are respectively represented as red, green, cyan, and blue nodes.
Also, to show how different embeddings for the same question are distributed, we randomly select 5\% of the questions and add colored lines that link their general question embeddings to their layerwise reference embeddings.
We observe that the reference embeddings of the three GNN layers are distributed in three distinct clusters positioned at approximately equidistant intervals, forming a roughly triangular pattern, which demonstrates the distinction of the reference embeddings in different layers.
It is interesting that the cluster of the general question embeddings is positioned near the center of the three reference embedding clusters despite a subset of nodes leaning towards the left side, which is in line with our intuition that the general question embeddings should present the general aspect of the questions.
Furthermore, to quantitatively demonstrate the clustering of general and layer-wise reference embeddings, we compute the Silhouette score~\citep{rousseeuw1987silhouettes} and the Calinski-Harabasz index~\citep{calinski1974dendrite} of these embeddings across the test questions in all six benchmarking datasets.
The Silhouette score quantifies how well-separated the clusters are, ranging from $[-1, 1]$, with values closer to 1 indicating better separation.
The Calinski-Harabasz index also measures clustering quality, considering within-cluster dispersion and between-cluster separation.
It ranges from $[0, \infty]$, where higher values imply better clustering.
The results, reported in \cref{table:clustering}, indicate that the general and layer-wise reference embeddings are well clustered for all benchmarking datasets, with Silhouette scores consistently greater than 0.6 and Calinski-Harabasz indices exceeding 1,400.

\begin{table}[]
\centering
\caption{The Silhouette score and Calinski-Harabasz index computed for quantifying the clustering of general and layer-wise question embeddings.}
\label{table:clustering}
\begin{tabular}{@{}ccc@{}}
\toprule
               & Silhouette Score & Calinski-Harabasz Index \\ \midrule
WQSP & 0.69             & 6,934.44                 \\
CWQ            & 0.71             & 8,800.87                 \\
PQ-2hop        & 0.73             & 2,049.99                 \\
PQL-2hop       & 0.85             & 1,722.30                 \\
PQ-3hop        & 0.84             & 6,461.70                 \\
PQL-3hop       & 0.76             & 1,496.35                 \\ \bottomrule
\end{tabular}
\end{table}

\begin{figure*}[t]
    \centering
    \includegraphics[trim=0pt 0pt 0pt 0pt, clip, width=1.\textwidth]{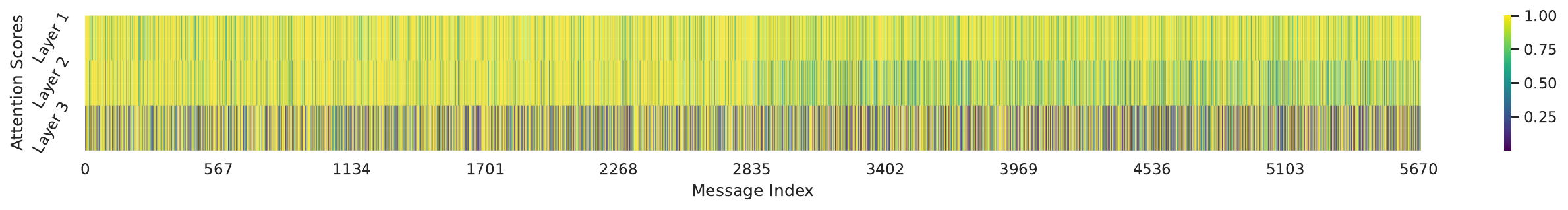}
    \caption{Distribution of the normalized message attention scores computed based on the layerwise reference embeddings in three GNN layers for answering the question ``\textit{what is the name of the place of birth of offspring of Henriette Adelaide of Savoy's husband?}''}
    \label{fig:heatmap}
\end{figure*}

As formulated in \cref{equ:individual_message_score}, the layerwise reference embeddings in each GNN layer are used to compute attention scores, which weigh the triples (specifically, messages of the triples) in the KG during message passing. 
Variations in the attention score distributions in different GNN layers indicate changes in the model's focus across layers.
In \cref{fig:heatmap}, we illustrate this using the case-study question, ``\textit{what is the name of the place of birth of offspring of Henriette Adelaide of Savoy's husband,}'' by visualizing the normalized distributions of attention scores computed in three GNN layers. 
The figure shows substantial changes in the distributions across layers, confirming that the model shifts its focus as it progresses.
Furthermore, we quantify these shifts by computing the average Jensen–Shannon (JS) divergence~\citep{DBLP:journals/tit/Lin91} between normalized attention score distributions from different GNN layers, evaluated over the test questions in six benchmarking datasets. 
The results are presented in \cref{table:js_divergence}, where the column ``Layers-1/2'' denotes the average JS divergence between the first and second layers' attention score distributions. 
JS divergence is bounded between 0 and 1, where 0 indicates identical distributions, and 1 indicates substantial divergence.
As shown in \cref{table:js_divergence}, JS divergence values around 0.5 are observed for complex questions such as those in WQSP and CWQ, indicating substantial shifts in the model's layer-wise focus. 
Conversely, the layer-wise divergence remains generally low for simpler questions such as PQ-2hop and PQL-2hop. 
This aligns with the intuition that simpler reasoning tasks require fewer shifts in focus across layers.

\begin{table}[]
\centering
\caption{The average Jensen–Shannon (JS) divergence between normalized distributions of message attention scores computed in different GNN layers.}
\label{table:js_divergence}
\begin{tabular}{@{}cclcc@{}}
\toprule
               & \multicolumn{2}{c}{Layers-1/2} & Layers-2/3 & Layers-1/3 \\ \midrule
WQSP & \multicolumn{2}{c}{0.66}      & 0.59      & 0.64      \\
CWQ            & \multicolumn{2}{c}{0.63}      & 0.46      & 0.66      \\
PQ-2hop        & \multicolumn{2}{c}{0.24}      & 0.17      & 0.08      \\
PQL-2hop       & \multicolumn{2}{c}{0.25}      & 0.20      & 0.10      \\
PQ-3hop        & \multicolumn{2}{c}{0.40}      & 0.66      & 0.60      \\
PQL-3hop       & \multicolumn{2}{c}{0.36}      & 0.30      & 0.15      \\ \bottomrule
\end{tabular}
\end{table}

\section{Discussion of Technical Contributions}

In this section, we discuss the technical contributions of our method by comparing it with existing KGQA methods and explaining how we achieved superior performance.
In addition, we compare our method with two similar but distinct methods that also jointly utilize KG structures and language models (LMs) for QA.\\
\textbf{1) Comparison with the previous SOTA methods---NSM and SR+NSM}. 
Similar to our proposed GNN model, both NSM and SR+NSM compute entity embeddings based on message propagation.
However, our GNN model is more sophisticated, with several differences and subsequent advantages.
First, our GNN model computes messages for each entity based on both the entity's neighboring relations and neighboring entities.
One layer of GNN encoding enables information exchange between entities one hop/triple away from each other.
Multiple layers of GNN encoding enable information exchange between entities multiple hops away from each other.
This is important for answering multi-hop questions, as their expected answers are multiple hops away from the topic entities mentioned in the questions.
However, in NSM and SR+NSM, the KG encoding essentially only learns the local representations for entities based on their surrounding relations.
Entities cannot exchange information by directly influencing each other's embeddings.
This severely limits the two models' performance in answering multi-hop questions.
Second, when updating entity embeddings, NSM and SR+NSM employ a feedforward neural network and directly concatenate the aggregation of received messages with the original embedding of each entity.
This is suboptimal as the aggregation of received messages is not necessarily useful and may need to be discarded regarding the given question.
In GNN2R, we propose an attentional gate that determines how much of the received messages should be added to the entity embedding with reference to the general question embedding.
Third, as introduced in \cref{sec:explainability_eval} (Comparison with the SOTA Methods), NSM and SR+NSM rely on the intermediate state of the KG encoding (i.e., intermediate entity probabilities) to derive reasoning chains.
However, the KG encoding is only trained with final answers.
There is a lack of supervision for the reasoning chains.
Consequently, the quality of generated reasoning chains of NSM and SR+NSM is very limited.
In contrast, we particularly propose a weakly supervised approach for generating reasoning subgraphs in GNN2R, which utilizes the structures of KGs to generate candidate reasoning subgraphs and leverages the knowledge of the pre-trained LM to select correct reasoning subgraphs.\\
\textbf{2) Comparison with TransferNet and earlier methods.}
As introduced in \cref{sec:explainability_eval} (Comparison with the SOTA Methods), TransferNet answers questions by passing scalar scores from topic entities to answer entities.
Compared with the high-dimensional embeddings that our GNN passes, the scalar scores lack the capability of encoding complex semantics, which limits the model's representation and reasoning capabilities.
Also, when computing score-passing probabilities for relations, TransferNet only considers the given question and ignores the semantics of the relations.
In contrast, the attentional message passing in our GNN fully leverages the semantics of relations by encoding them into the messages.
Regarding the generation of reasoning chains, TransferNet has the same weakness as NSM and SR+NSM: the intermediate reasoning states of the models not trained with reasoning subgraph annotations are highly unreliable.
In contrast, we propose a weakly-supervised approach for fine-tuning the LM and propose an effective method (i.e., Step-II) to leverage the pre-trained knowledge and language understanding capabilities of the LM in GNN2R.
Regarding the earlier baselines, including Relation Learning, GraftNet, and EmbedKGQA, the major difference is that GNN2R effectively leverages the strengths of both the GNN (KG embeddings) and the LM.
These baselines either only focus on LMs (e.g., Relation Learning) or only focus on KG embeddings (e.g., GraftNet and EmbedKGQA).
Also, the weakly-supervised retrieval of reasoning subgraphs in GNN2R is not considered in these methods.\\
\textbf{3) Comparison with relevant KG structure+LM QA methods.}
The framework of LM+GNN has been recently studied in other KGQA domains~\citep{DBLP:conf/naacl/YasunagaRBLL21,DBLP:journals/corr/abs-2201-08860}.
QA-GNN~\citep{DBLP:conf/naacl/YasunagaRBLL21} and GreaseLM~\citep{DBLP:journals/corr/abs-2201-08860} are two similar methods that answer multiple-choice questions based on KGs.
The similarities between our method and the two methods are two-fold:
1) LMs are commonly leveraged to capture the semantics of textual questions, including the semantic similarity comparison in our method and the relevance score computation in QA-GNN.
2) We all employ GNN modules to reason over the graph structures of KGs, including the GNN-based coarse reasoning in our method and the GNN layers in GreaseLM.
However, as we focus on a different task setup, our method is inherently different regarding the proposed GNN structure and the application of LMs.
Specifically, QA-GNN and GreaseLM are provided with a very limited set of answer choices for each question.
They can perform fine-grained reasoning for each answer choice, explicitly computing its probability of being correct.
In our task, all entities in the underlying KG have to be examined, which is more challenging and prohibits fine-grained reasoning regarding specific entities.
Therefore, our GNN module aims to concurrently represent and rank all entities in the embedding space.
Another major difference is that LMs are more integrated into the GNN reasoning of the two methods.
Especially in GreaseLM, the modality interaction layer enables bidirectional information exchange between the LM and GNN layers.
In our method, we only employ the LM after efficiently pruning the search space of answers and reasoning subgraphs.

\section{Limitations and Future Work}
\label{sec:limitations}

In the following, we discuss GNN2R's limitations and the future work:

\textbf{Limitations.} 1) As a retrieval-based method, GNN2R excels in retrieving answer entities but lacks the capabilities of post-processing and aggregating answer entities for questions that require counting, sorting, filtering, comparing, and set operations.
2) In Step-II, GNN2R generates candidate reasoning subgraphs by extracting paths between topic entities and candidate answers.
It could be challenging to handle KGs containing extensive auxiliary nodes (e.g., blank nodes) that require adapting the path extraction to the specific KG schema.
3) Another limitation of GNN2R is that the GNN encoding is not directly scalable to very large KGs.
As discussed in \cref{subsection:step_1} (Scalability Discussion), GNN2R requires preparing question-relevant subgraphs that cover all question-relevant triples but remain a manageable size.
4) The initialization of entity/relation embeddings in the GNN encoding is based on the entity/relation labels in the KG. 
Therefore, regarding the quality requirement of KGs, one major limitation of GNN2R is that it requires semantically meaningful labels.

\textbf{Future Work.} 1) We intend to improve the model's capability of answering more complex questions, especially the questions that require complex operations over the retrieved answers.
The plan is to combine GNN2R---a retrieval-based approach---with logical query-based approaches, such as adding a module that can generate specific SPARQL queries for post-processing.
2) We also plan to further reduce the model's dependency on training data.
For the GNN module, we intend to investigate if the amount of required training data can be minimized via transfer learning, such as training the GNN on a collection of QA datasets and applying it to new datasets with very limited training.
For the LM, we intend to explore if the recent LLMs can be prompted to compare questions with reasoning subgraphs in a zero-shot or few-shot setup.

\section{Conclusions}

In this paper, we propose a novel KGQA method called GNN2R that can efficiently retrieve both final answers and reasoning subgraphs as answer explanations in two steps:
In \emph{Step-I GNN-based coarse reasoning}, a novel GNN architecture is proposed to represent the given question and KG entities in a joint embedding space, where the question is close to its answers.
This step efficiently prunes the search space of candidate answers and reasoning subgraphs, ensuring the high efficiency of GNN2R.
In \emph{Step-II LM-based explicit reasoning}, we propose to jointly utilize the KG structures and the language understanding capability of a pre-trained LM to extract reasoning subgraphs.
We accordingly propose algorithms to fine-tune the LM with only weak supervision, i.e., question-final answer pairs.
Extensive experiments have been conducted on multiple benchmark datasets to evaluate the proposed method in comparison with recent SOTA methods, which demonstrate the superior performance of GNN2R regarding effectiveness, efficiency, and the quality of generated reasoning subgraphs. 
We have also conducted extensive experiments with in-depth analysis to evaluate the robustness of GNN2R and the contributions of its included modules.

\section*{Acknowledgments}
This work was partially funded by the University Research Priority Program ``Dynamics of Healthy Aging'' at the University of Zurich, 
the Swiss National Science Foundation through project MediaGraph (No. 202125), 
and the Horizon Europe Graphmassivizer project (No. 101093202). 

\bibliography{bibfile}

\end{document}